\documentclass[12pt]{article}

\usepackage[utf8]{inputenc} 
\usepackage{amsmath} 
\usepackage{graphicx} 
\usepackage{hyperref} 
\usepackage[left=3cm,right=3cm,top=2cm,bottom=2cm]{geometry} 
\usepackage{authblk} 
\usepackage[round]{natbib}
\usepackage{appendix}
\usepackage{changepage} 

\usepackage{tabularx}
\usepackage{longtable}
\usepackage{booktabs} 
\usepackage{multicol} 
\usepackage{multirow} 
\usepackage{ltablex} 
\usepackage{caption} 
\usepackage{pdflscape}
\usepackage{afterpage}
\usepackage{tablefootnote}
\usepackage[T1]{fontenc}
\usepackage{multirow}
\usepackage{graphicx} 

\usepackage{xcolor}
\usepackage{amssymb}

\usepackage{float} 
\usepackage{placeins} 

\newcommand{\appendixnumbering}{%
    \setcounter{figure}{0}%
    \renewcommand{\thefigure}{Appendix~\Alph{section}}%
    \setcounter{table}{0}%
    \renewcommand{\thetable}{Appendix~\Alph{section}}%
}

\newenvironment{acknowledgment}%
  {\clearpage\thispagestyle{empty}\null\vfill\begin{center}%
  \bfseries Acknowledgment\end{center}}%
  {\vfill\null}

\usepackage{soul}

\usepackage{etoolbox} 

\newcommand{\keywords}[1]{\par\noindent\textbf{Keywords:} \textit{#1}}

\usepackage{fancyhdr} 
\title{Annotating Scientific Uncertainty: A comprehensive model using linguistic patterns and comparison with existing approaches}
\author[1]{Panggih Kusuma Ningrum\thanks{Corresponding author: panggih\_kusuma.ningrum@univfcomte.fr}}
\author[2]{Philipp Mayr}
\author[2]{Nina Smirnova}
\author[1,3]{Iana Atanassova}
\affil[1]{Université Marie et Louis Pasteur, CRIT, F-25000 Besançon, France}
\affil[2]{GESIS –- Leibniz Institute for the Social Sciences, Cologne, Germany}
\affil[3]{Institut Universitaire de France (IUF), France}

\date{} 

\begin{document}


\maketitle 

\begin{abstract}
We present UnScientify\footnote{\url{https://bit.ly/unscientify-demo}}, a system designed to detect scientific uncertainty in scholarly full text. The system utilizes a weakly supervised technique to identify verbally expressed uncertainty in scientific texts and their authorial references. The core methodology of UnScientify is based on a multi-faceted pipeline that integrates span pattern matching, complex sentence analysis and author reference checking. This approach streamlines the labeling and annotation processes essential for identifying scientific uncertainty, covering a variety of uncertainty expression types to support diverse applications including information retrieval, text mining and scientific document processing. The evaluation results highlight the trade-offs between modern large language models (LLMs) and the UnScientify system. UnScientify, which employs more traditional techniques, achieved superior performance in the scientific uncertainty detection task, attaining an accuracy score of 0.808. This finding underscores the continued relevance and efficiency of UnScientify's simple rule-based and pattern matching strategy for this specific application. The results demonstrate that in scenarios where resource efficiency, interpretability, and domain-specific adaptability are critical, traditional methods can still offer significant advantages. 
\end{abstract}

\keywords{rule-based, pattern matching, research article, uncertainty, LLM}

\section{Introduction}
Uncertainty is an inherent part of scientific research, as the very nature of scientific inquiry involves posing questions, developing hypotheses, and testing them using empirical evidence. Despite the best efforts of scientists to control for extraneous variables and obtain accurate measurements, there is always a certain degree of uncertainty associated with any scientific findings. This uncertainty can arise from a variety of sources, such as measurement error, sampling bias, or limitations in experimental design. Consequently, researchers resort to various strategies to manage and mitigate uncertainty when presenting their findings in academic articles. These may include using language that is overly definitive or hedging their claims with qualifiers such as "presumably" or "possible" \citep{hyland_talking_1996}.

Uncertainty is a complex concept with various definitions (see, e.g. \cite{walker_defining_2003, Refsgaard,ASCOUGH2008383}). For instance, \cite{SIGEL2010502} defined uncertainty in the scope of water management as \textit{"Lack [of] confidence about knowledge relating to a specific question"}, while \cite{lehmann} interpreted uncertainty in environmental research as a \textit{"Measure of unexplained variation"}.
The literature offers a wide range of meanings and interpretations of the term such as lack of knowledge, doubt, inconsistency, ambiguity, vagueness, subjectivity, error, risk, ignorance, and contradiction \citep{popper_karl_2002, ruhrmann_frames_2013, schwandt_credible_2015, klir_fuzzy_1988}. The terminology used in describing scientific uncertainty varies depending on its nature and characteristics, disciplines, sources, and elements related to the individuals involved in scientific activity resulting plethora of meanings and connotations. When one uses the term, the context shapes the specific meaning and connotation used.

Journal articles are a common medium used by scientists to communicate their ideas and research results. As such, the textual content of these articles is a valuable source for studying scientific uncertainty. They are subjected to extensive independent peer review, which allows to guarantee the validity of the research methods and results that are reported. Most importantly, journal articles play an important role in disseminating knowledge to a wider audience. They not only describe the structural thinking of the author(s), depict the author’s persona, and explain the research and analysis process \citep{candlin2014writing,hyland_talking_1996, Candlin2000,Hyland2000DevelopmentsIE}. Furthermore, journal articles invite to discuss ideas and research results, which are not always unequivocally clear but may involve some uncertainties that give space for further improvement and exploration.

Given the wide variety of definitions and scope used to describe uncertainty, it is not surprising that significant efforts have been made to detect and identify uncertainty in science. Previous research has mainly focused on identifying a specific set of uncertainty cues or markers, and text segments that express uncertainty in scholarly articles, using the abstract \citep{vincze_bioscope_2008} or the full text \citep{medlock_weakly_2007, atanassova_studying_2018, riccioni_self-mention_2021}. These studies have helped expand the vocabulary and lexicon associated with uncertainty. However, their practical application is often ineffective because of the intricate nature of natural language.

The task of identifying scientific uncertainties remains challenging due to several factors. Firstly, there is a scarcity of extensively annotated corpora that cover the full spectrum of scientific uncertainty. Existing datasets, such as the BioScope corpus \citep{szarvas_bioscope_2008}, focus narrowly on negation and speculation in biological abstracts, while the FACTBANK corpus \citep{sauri_factbank_2009} targets the veracity of event mentions. Similarly, the Genia Event corpus \citep{kim_corpus_2008} is limited to annotating biological events with negation. These corpora, along with others focusing on hedging, modality, hypothetical markers, or weasel words \citep{riccioni_self-mention_2021, medlock_weakly_2007, ganter-strube-2009-finding, szarvas_cross-genre_2012, styler_iv_temporal_2014}, are restricted to specific types of general uncertainty and fail to capture the diversity of scientific uncertainty expressions across domains.

Scientific uncertainty, however, encompasses a broader range of expressions, including research limitations, variability in findings, and ungeneralizable conclusions, which may not rely on traditional cues like hedging or modality \citep{Ningrum2023}. These expressions are often overlooked in existing datasets, limiting their applicability to the broader scope of scientific uncertainty addressed in this study. Consequently, there is an urgent need for more diverse corpora, covering a wider range of uncertainty types and domains, to enable a more comprehensive understanding of uncertainty in natural language processing (NLP). This gap reinforces the importance of developing a robust framework and annotated dataset to advance the detection and analysis of scientific uncertainty.

Secondly, identifying expressions of uncertainty is inherently challenging due to the ambiguity of linguistic cues and the diverse ways uncertainty can be expressed, depending on the writing style or stance of the scientist. A significant obstacle lies in the polysemy of cues and markers—words or phrases that carry multiple meanings depending on their context. This polysemy complicates uncertainty detection, as the same cues can convey entirely different meanings in different contexts. For instance, \cite{suhadi2017epistemic} explored the phenomenon of "two sides of the same coin modality" highlighting how the same modality word can imply distinct meanings, not all of which convey genuine uncertainty.

Another challenge concerns scientists' discourse in scientific writing. A typical scientific text contains various statements and information that discuss not only the present study but also former studies \citep{stocking_constructing_1993}. Scientists can use uncertainty claims from other studies as a rhetorical tool to persuade or describe and organize the state of knowledge. As a result, distinguishing the reference of the uncertainty feature -- whether the statement demonstrates uncertainty in the current study or former studies, is a crucial factor in better understanding the context of scientific uncertainty. A study by \cite{bongelli_writers_2019} is one of the few that was aware of this concern. In more detail, this study only focused on the certainty and uncertainty expressed by the speakers/writers in the here-and-now of communication and excluded those that were expressed by the other party.


\subsection{Related Works}
A wide range of techniques has been explored to detect and analyze uncertainty or related phenomena such as hedging, modality, and factuality in text. These techniques generally focus on two main objectives: treating uncertainty detection as a sentence-level classification task, where the text is categorized as either expressing uncertainty or not, and identifying specific linguistic cues or markers of uncertainty. The approaches used in previous studies can be grouped into three categories: rule-based methods, machine learning approaches, and deep learning or transformer-based methods. Each offers unique advantages and faces distinct limitations.

\subsubsection{Rule-Based Approaches}

Rule-based techniques represent some of the earliest efforts to detect uncertainty in text, relying on predefined cues, markers, or linguistic patterns. These approaches have typically employed handcrafted lexicons to identify linguistic indicators of uncertainty. For example, early works by \cite{MURDUENAS2021103131} employed such lexicons to detect hedges in English and Spanish research articles, while more recent studies, such as \cite{10.1007/s11192-023-04759-6}, leveraged  Mur-Dueñas' academic hedging lexicon to map hedging cues in scientific articles, analyzing the diachronic development of uncertainty expressions. However, as noted by \citet{Ningrum2023}, cues alone are not always reliable indicators of uncertainty, as their contextual meaning can vary significantly, making simple cue matching insufficient.

To address these limitations, researchers have advanced rule-based methods by incorporating syntactic patterns and regular expressions (RegEx) to capture more complex linguistic structures. For instance, \citet{chapman-etal-2007-context} introduced ConText, a system that combines cue detection with "pseudo-triggers" to disambiguate uncertain assertions. Similarly, \citet{atanassova_studying_2018} analyzed hedge verbs and an extended vocabulary of hedging cues to study uncertainty in biomedicine and physics, while \citet{rey:tel-03780719} explored uncertainty in climate change discussions, organizing uncertainties into reasoning types (abductive, inductive, deductive) and linking them to quantitative references. Despite these advancements, the rigid nature of regular expressions limit these approaches in capturing the flexibility and variability inherent in natural language. Additionally, these methods often remain confined to specific domains and contexts, reducing their generalizability.

Further refinements in rule-based methods include leveraging linguistic insights to tag nuanced and complex expressions. \citet{omero_writers_2020} implemented a system to identify if-clauses expressing uncertainty in biomedical scientific articles. Although their approach demonstrated the value of linguistic features, it struggled with capturing if-clauses involving more intricate patterns.

Similarly, \citet{muller_communicating_2021} employed a rule-based approach incorporating manual annotation, frame-linguistic modeling, and corpus-linguistic operationalization to detect uncertainty expressions in British and German newspaper articles about the coronavirus pandemic. They combined manual annotation with pattern-based searches, known as the “annotation-by-query” technique, to identify semantic concepts of social uncertainty. Building on this study, \citet{muller_corpus_2023} developed the application Sketch to detect expressions of ignorance and uncertainty specifically in political science texts and conspiracy theory texts. \citet{muller_corpus_2023} applied this technique to annotate the Darmstadt International Relations Corpus \citep{muller_darmstadt_2020}, the conspiracy theory subcorpus of LOCO \citep{miani_loco_2022}, and a reference corpus consisting of randomly selected articles from US national newspapers. The results indicate that this technique offers the advantage of being applicable to much larger datasets than purely qualitative annotation approaches, while also allowing efficient control over textual appearances and potential ambiguity through accessible concordances. However, a significant limitation of this approach lies in its ability to address the nuanced meaning of uncertainty expressions within complex sentences. Such sentences often contain uncertainty cues or markers along with elements of rebuttal such as neutrality, negation, or confirmation, which can shift the overall meaning of the sentence (see Section \ref{workflow} in the Complex Sentence Checking part for a detailed explanation). This complexity poses a challenge for the existing pattern-based and rule-based detection methods, as they often fail to account for the broader context necessary to accurately identify uncertainty expressions. As a result, these methods risk misinterpreting or overlooking genuine uncertainty, particularly in sentences where multiple linguistic elements interact to alter the intended meaning. This limitation underscores the need for more advanced approaches that incorporate contextual and semantic analysis to improve the accuracy of uncertainty detection in scientific texts.

\subsubsection{Machine Learning Approaches}

Machine learning (ML) methods have significantly advanced uncertainty detection by automating token classification and sequence labeling tasks. These methods were prominent in initiatives such as the CoNLL-2010 Shared Task \citep{morante-daelemans-2009-learning, clausen-2010-hedgehunter, rei-briscoe-2010-combining, Fernandes, mamani-sanchez-etal-2010-exploiting, li-etal-2010-exploiting, tang-etal-2010-cascade, zhang-etal-2010-hedge}, where models identified and labeled tokens expressing uncertainty. ML-based approaches rely on feature-rich representations that include surface-level attributes, part-of-speech (POS) tags, and syntactic relations. Some studies extended these feature sets to incorporate dependency relations derived from uncertainty cues, as demonstrated by \citet{rei-briscoe-2010-combining}, \citet{mamani-sanchez-etal-2010-exploiting}, and \citet{zhang-etal-2010-hedge}.

Some studies also employed weakly supervised methods to further refined their machine learning models by reducing reliance on extensive labeled datasets. For instance, \citet{medlock_weakly_2007} and \citet{szarvas_bioscope_2008} annotated six papers to create a test set and randomly selected 300,000 sentences as training data for iterative bootstrapping. These techniques used minimal seed annotations to iteratively expand training datasets, effectively leveraging linguistic features such as span detectors to capture uncertainty patterns. 

Despite these advancements, ML methods are not without limitations. Findings from the CoNLL-2010 Shared Task \citep{farkas-etal-2010-conll} revealed that even with the inclusion of complex linguistic feature sets, performance improvements were minimal. These results underscore the critical need for effective cue disambiguation and emphasize the importance of further research into domain adaptation and the development of models capable of handling uncertainty across diverse contexts.

One major limitation lies in the reliance on handcrafted features and predefined cues, which often fail to capture the nuanced semantics and syntactic intricacies of uncertainty expressions. Narrowly focused datasets, as seen in prior studies targeting phenomena like hedging or speculative assertions, constrain models and impede their ability to generalize to different domains or capture broader types of uncertainty. For example, Medlock and Briscoe's (2007) model struggled to identify assertive statements of knowledge paucity, as these statements did not rely on explicit cues included in the training set. 

Moreover, \citet{omero_writers_2020} highlighted the difficulty ML models face in detecting complex uncertainty assertions, particularly those embedded in syntactically intricate constructions like if-clauses. This challenge may stem from limitations in how features are encoded during training. The reliance on fixed cue spans restricts ML models from addressing long-distance dependencies and intricate word interactions, both of which are crucial for capturing complex uncertainty expressions. These constraints highlight the need for methods that integrate deeper semantic and syntactic understanding to improve the detection of nuanced uncertainty patterns.

\subsubsection{Large Language Models and Transformer-Based Approaches}

The advent of deep learning and transformer-based models has revolutionized text classification including speculation assertion and negation detection by enabling the integration of contextual embeddings and word representations. Models like BERT, XLNet, and RoBERTa have demonstrated significant advancements in detecting uncertainty markers. \citet{khandelwal-britto-2020-multitask} evaluated these models on datasets such as the BioScope Corpus \citep{szarvas_bioscope_2008} and the SFU Review Corpus \citep{Kolhatkar2020-pw} for the detection of speculation and negation, leveraging contextual embeddings for greater precision. They proposed a joint training approach for cue detection and scope resolution, which outperformed separately trained versions of the same transformer-based architectures. This method achieved new state-of-the-art results, with the Combined Early Stopping variant delivering the best performance, highlighting the effectiveness of these models in capturing complex speculation and negation expressions.

The models' ability to leverage contextual embeddings and word representations marks a significant leap in speculation and negation detection, therefore offer exciting potential for further exploration, particularly in addressing diverse and nuanced types of scientific uncertainty. However, the conceptual boundaries of their training datasets, often based on simplified classifications, present an opportunity to expand their applicability. By experimenting with richer and more varied annotations, these models may potentially capture the complexity of scientific uncertainty expressions more effectively.

\subsection{Research Objectives}

Based on the rationales and research problems discussed in previous sections, the primary objective of our study is to develop a system for automatically annotating sentences that express scientific uncertainty in scholarly articles. The system goes beyond basic cue detection by incorporating deeper context and semantic analysis, ensuring the system not only identifies uncertainty but also captures its contextual and semantic nuances.

To achieve this, our study builds upon the uncertainty annotation framework presented in \cite{Ningrum2024-wi}. We introduce UnScientify, a weakly supervised system that combines a fine-grained annotation framework, pattern matching, linguistic features, and complex sentence analysis to detect scientific uncertainty in the full text of articles across diverse domains. UnScientify also extracts deeper contextual information, such as authorial reference, distinguishing between uncertainty expressed by the authors and uncertainty reported from cited studies. This focus on authorial reference addresses a gap in prior research, offering a nuanced understanding of how uncertainty is framed and attributed in scientific discourse. By prioritizing interpretability and minimizing the need for fully annotated datasets, UnScientify provides a comprehensive and efficient solution for analyzing scientific uncertainty in scholarly articles.


To summarise, we focus on two main research objectives:

\begin{itemize} 
    \item \textbf{Develop an Automated System for Scientific Uncertainty Detection}: Build a system to detect and annotate scientific uncertainty in scholarly articles across various domains with interpretable results. \item \textbf{Provide a Publicly Available Dataset}: Create and release a gold-standard annotated corpus for scientific uncertainty, serving as a benchmark for future research and model evaluation.
\end{itemize}

The current study represents a substantially extended version of the article presented at the Joint Workshop of the 4th Extraction and Evaluation of Knowledge Entities from Scientific Documents and the 3rd AI + Informetrics (EEKE-AII2023) \citep{eeke_2023}, and the authors appreciate the insights gained from the workshop that contributed to the refinement of the research.

\section{Materials and Methods}

\subsection{Conceptual Definition of Scientific Uncertainty}

In order to set a clear scope for the current study, we will focus on the uncertainty expressed in the scientific articles. After studying the different definitions of uncertainty in the previous works cited above, we provide a conceptual definition of scientific uncertainty that will be used in the current study. In particular, we need to be able to distinguish the sentences in articles (or other text spans) that express uncertainty from those that do not express uncertainty. We define them as follows:

\begin{quote}
    \textit{A sentence expresses uncertainty if it (explicitly) expresses a state of absence of knowledge, insufficient knowledge, or lack of precision in information about a concept or subject.}
\end{quote}

\subsection{Data}

The present study employs three annotated corpora. These corpora consist of 59 journals from four different disciplines: Medicine; Biochemistry, Genetics \& Molecular Biology; Interdisciplinary \& miscellaneous; and Empirical Social Science\footnote{All social science articles are from SSOAR (\url{https://www.ssoar.info/}); we selected articles from 53 social science journals indexed in SSOAR.} which represent Science, Technology, and Medicine (STM) as well as Social Sciences and Humanities (SSH). The corpora consist of 975 randomly selected English sentences from 312 articles across 59 journals. These sentences were annotated to identify uncertainty expressions and authorial references. Using multiple corpora from different disciplines, this study aims to capture a diverse range of uncertainty expressions and improve the generalizability of the results. Table \ref{table:corpora} illustrates the distribution of the data in the corpora and Table \ref{table:sample_sentences} shows the sample of annotated sentences.

\begin{table}[htbp]
    \centering
    \footnotesize
    \caption{Corpora description}
    \label{table:corpora}
    \begin{tabularx}{\textwidth}{llrr}
        \toprule
        Discipline & Journal & Articles & Sentences \\
        \midrule
        Medicine & a. BMC Med & 51 & 95 \\
         & b. Cell Mol Gastroenterol Hepatol & 25 & 36 \\
        \multirow[t]{100}{*}{\parbox{5.7cm}{\vspace{1.3\baselineskip}Biochemistry, Genetics \& Molecular Biology}} & a. Nucleic Acids Res & 52 & 63 \\
         & b. Cell Rep Med & 22 & 48 \\
        Interdisciplinary \& miscellaneous & a. Nature & 34 & 57 \\
         & b. PLoS One & 42 & 55 \\
        Empirical Social Science & SSOAR (53 journals) & 86 & 621 \\
        \midrule
        \multicolumn{2}{l}{{Total}} & 312 & 975 \\
        \bottomrule
    \end{tabularx}
\end{table}

\addtocounter{table}{-1} 

\begin{table}[h]
    \centering
    \footnotesize
    \caption{Sample of annotated sentences}
    \label{table:sample_sentences}
    \begin{tabularx}{\textwidth}{Xll}
        \toprule
        Sentence & Uncertainty & Authorial Ref. \\
        \midrule
        It is possible that corticosteroids prevent some acute gastrointestinal complications. & Yes & Author(s) \\
        In this test, a likelihood ratio test statistic is calculated for the two tree versus one tree models, and compared to a null distribution generated by non-parametric bootstrapping (see Methods). & No & - \\
        Previous meta-analyses have shown a significant benefit for NaHCO3 in comparison to normal saline (NS) infusion [6,7], although they highlighted the possibility of publication bias. & Yes & Former/Prev. Study(s) \\
        \bottomrule
    \end{tabularx}
\end{table}

Two sampling methods were implemented to produce the dataset: cues mapping and manual searching. The first method used lists of uncertainty cues and markers from \citet{hyland_talking_1996}, \citet{chen_scalable_2018}, and \citet{bongelli_writers_2019} (see \ref{tab:Cues_list}). These cues were mapped through our corpora using Regular Expression, and from the sentences detected with these cues, we randomly selected 200 sentences from each journal in Medicine; Biochemistry, Genetics \& Molecular Biology; and Interdisciplinary \& miscellaneous disciplines, and up to 900 sentences from the Empirical Social Science corpus, resulting in 2100 sentences. After removing duplicates, 880 sentences remained.

The second method involved randomly selecting two articles from each journal in Medicine; Biochemistry, Genetics \& Molecular Biology; and Interdisciplinary \& miscellaneous disciplines. This manual search accounted for the flexibility of natural language, recognizing that uncertainty expressions may vary and not strictly follow predetermined cues. This step ensured the inclusivity of potential variations, enhancing the robustness of our analysis. From this method, 95 data points were obtained, resulting in a final dataset of 975.

Manual annotation was then conducted on the dataset. First, each sentence was labeled for the presence of a Scientific Uncertainty (SU) expression (Yes/No). For sentences marked "Yes," the annotators identified the SU markers (span) and the authorial reference, labeling it as 1: author(s), 2: former study(ies), or 3: both author(s) \& former study(ies). Two annotators, provided with explicit guidelines (see \ref{fig:codebook}) and reference text data, were trained and tested together to ensure consistency. They then independently labeled the dataset, resolving any discrepancies through discussion and consensus. In rare cases of disagreement, a third annotator made the final decision. This corpora construction process adheres to the framework outlined by \citet{Ningrum2024-wi}. The annotated dataset used in this study is publicly accessible on the Zenodo platform \footnote{AURORA-MESS: Annotated Uncertainty and Reference in Open Research Articles for Multidisciplinary and Empirical Social Science: \url{https://zenodo.org/records/15001250}} \citep{Ningrum2025-mb}.

\subsection{Approach}
Identifying scientific uncertainty in academic texts is a complex task due to various reasons. Previous research indicates that relying solely on cues or markers such as hedging words or modal verbs may not accurately detect and identify scientific uncertainty \citep{Ningrum2023}. The natural language and writing styles used by scientists, along with variations in domain-specific terminology, add to the complexity of identifying uncertainty in scientific text. To address these limitations, our research proposes a fine-grained annotation scheme for identifying uncertainty in scientific texts and their authorial reference.

\subsubsection{Fine-grained SU annotation scheme and patterns formulation}
The current study adopts a span-based method to identify scientific uncertainty in academic texts, building on prior research indicating that word-long phrases enhance the detection of hedging and uncertainty  \citep{szarvas_bioscope_2008, Ningrum2023}. This approach goes beyond mere linguistic cues, incorporating analysis of linguistic patterns and features such as Part-of-speech (POS) tags, morphology, and dependency within the spans.

A list of annotated spans has been created and classified into twelve groups of scientific uncertainty (SU) patterns based on their semantic meaning and characteristics. The groups include conditional expressions, hypotheses, predictions, and subjectivity, among others. In other words, the classification is based on the types of expressions used to convey uncertainty and the context in which they are used.

Additionally, the scheme considers spans of text that signal disagreement statements as one of SU groups, despite ongoing debate regarding whether disagreement expressions should be considered as such. The justification for this approach is rooted in the idea that uncertainty in research can stem from conflicting information or data, where multiple sources provide contradictory knowledge \citep{zimmermann_application-oriented_2000}. This type of uncertainty cannot be reduced by increasing the amount of information.

\begin{figure}[!hbt]
  \centering\includegraphics[width=0.5\columnwidth]{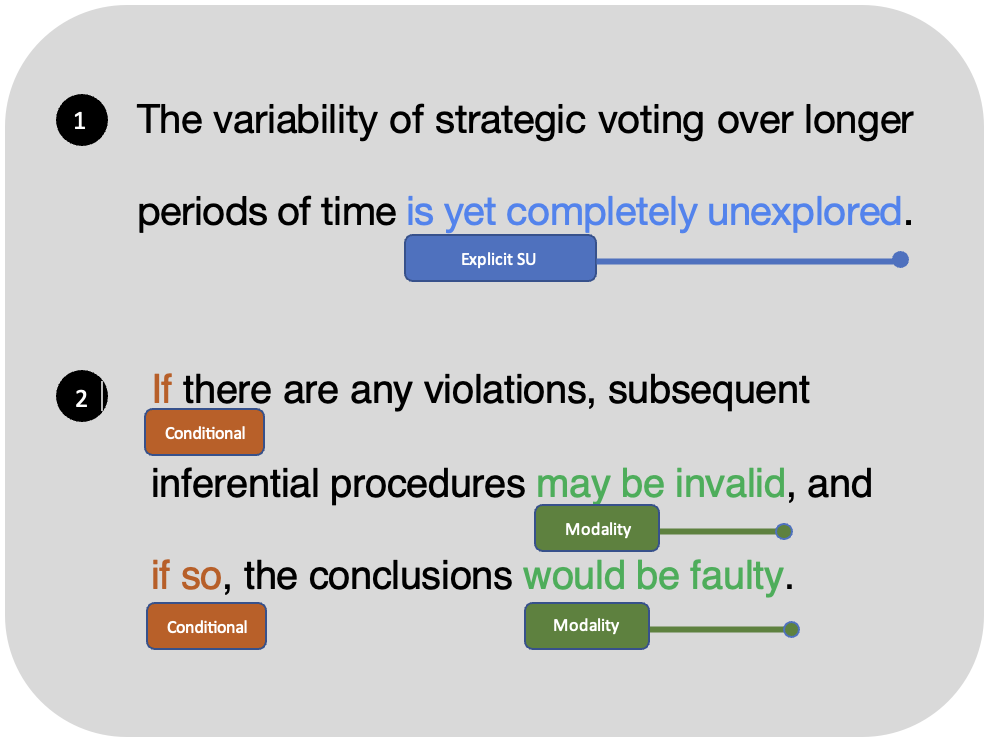}
  \caption{Two annotated sentences with SU expressions. Samples of output from the span annotation process are shown in different colours based on their SU Pattern Group.}
  \label{fig:Span_labelling}
\end{figure}

Once the annotated spans are classified, Scientific Uncertainty Span Patterns (SUSP) are formulated based on the word patterns of each span and its linguistic features. Figure \ref{fig:Span_labelling} shows examples of the output from the spans annotation process. Each span is assigned a label corresponding to its SU pattern group. It should be noted that a sentence can have multiple labels assigned to different SU pattern groups, as seen in the second example, where labels for both conditional expression and modality are present. This feature of our annotation scheme enables the identification of complex expressions of uncertainty in scientific text. Table \ref{tab-examples} shows more details about the list of SU pattern groups and samples from each group. More information about the pattern formulation process can be seen in \ref{fig:patterns_formulation}.

\subsubsection{Authorial Reference Checking}
Authorial reference is crucial in scientific writing to provide context, especially when identifying scientific uncertainty. It helps to indicate the authorship of the argument and distinguish between the claims of the author and those of others. This can be achieved through various styles of authorial reference, such as in-text citations, reference or co-reference \citep{powley_evidence-based_2007}. Additionally, there are disciplinary variations in both the frequency and use of personal and impersonal authorial references \citep{khedri_how_2020}.
Proper attribution of uncertain claims is important to determine their origin and evaluate the credibility of the argument. For instance, when stating a hypothesis, it is essential to indicate whether it is the author's hypothesis or cited from another source. This helps the reader to assess the level of uncertainty associated with the statement. 

In the present study, the authorial reference of each sentence was annotated based on the citation and co-citation patterns, and the use of personal and impersonal authorial references. Furthermore, sentences were labeled into three groups including 1) author(s) of the present article, or 2) author(s) of previous research. The last group, 3) both, is intended to accommodate complex sentences in which uncertainty may refer to both the author(s) and the previous study(s). Here, we present some examples of typical authorial reference mentions in context:
\begin{quote}
    1. $<\text{I/We/Our study...}> <\text{present(s)/introduce(s)/contribute(s) to}> ...$ \\
    2. $<Author/The former study...> <text>$ \\
    3. $(Author) (Year) <Text>$ \\
    4. $<Text> (Author1, Year1; Author2, Year2 . . .)$ \\
    5. $<Text> [Ref-No1, Ref-No2 . . . ]$
\end{quote}

\subsection{UnScientify Workflow} \label{workflow}

The UnScientify demo system for identifying scientific uncertainty expressions in articles is built using the spaCy framework. We decided to use the spaCy python library for its robust linguistic analysis capabilities, including extraction of features such as part-of-speech (POS) tags, lemmas, morphology, and syntactic dependencies, along with its efficient processing pipeline. Its extensible pattern matcher framework enables seamless integration of expert-driven rules, aligning well with our weakly supervised approach. These features make spaCy a practical choice for rule-based methods and reproducible research.
While we acknowledge that spaCy’s default models may lack the deep contextual representations, such as those of transformer-based models, its lightweight architecture, focus on linguistic features, and compatibility with our task-specific requirements proved highly advantageous.

\begin{figure*}[htbp]
    \centering
    \includegraphics[width=\textwidth]{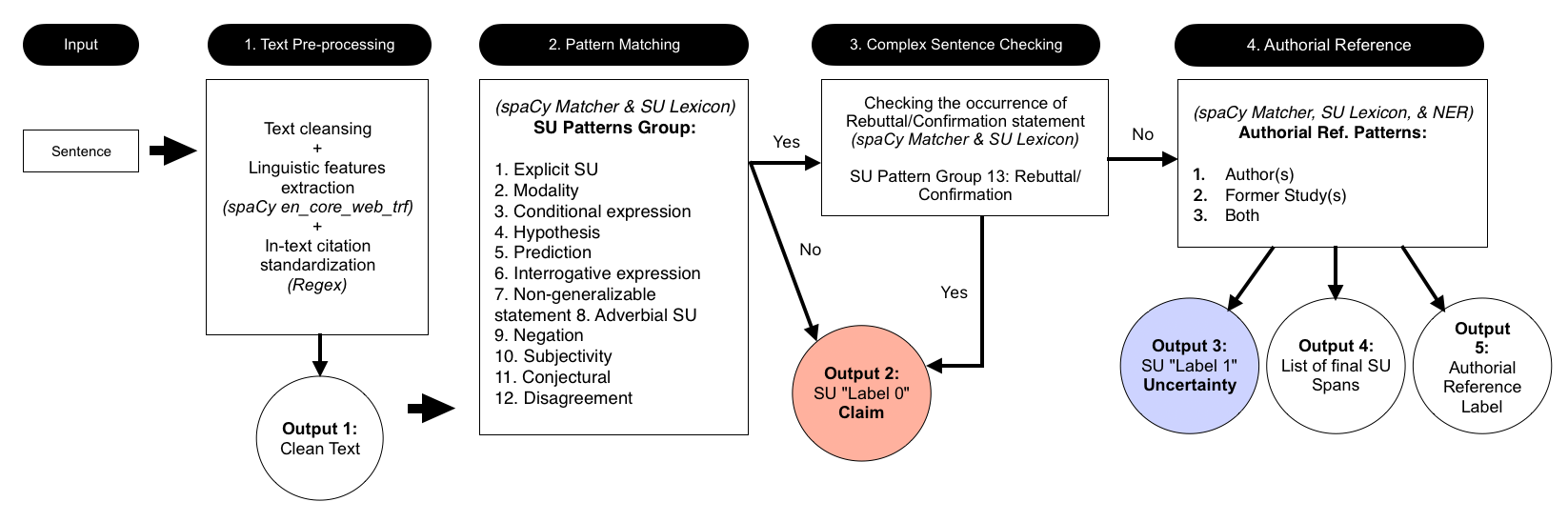}
    \caption{Scientific Uncertainty (SU) expression identification workflow}
    \label{fig:SU_iden_flow}
\end{figure*}

Figure \ref{fig:SU_iden_flow} shows the workflow of UnScientify. It consists of four main components: 1) Text Pre-processing, 2) Pattern Matching, 3) Complex Sentence Checking, and 4) Authorial Reference Checking.

The first step, Text Preprocessing, is designed to ensure the integrity and clarity of the data prior to analysis. This critical phase includes data cleansing, which entails removing irrelevant elements from text that might obscure or distort the analysis of scientific uncertainty. Furthermore, linguistic features such as part-of-speech (POS) tags, morphology, and dependency information are extracted from the input text. In this step, the \texttt{en\_core\_web\_trf} from spaCy is used in UnScientify system (see \ref{table:spaCy_models}). The selection of the \texttt{en\_core\_web\_trf} pipeline was driven by its superior performance across key linguistic tasks essential for our system, outperforming other spaCy pipelines. Unlike the tok2vec-based pipelines in \texttt{en\_core\_web\_sm}, \texttt{en\_core\_web\_md}, and \texttt{en\_core\_web\_lg}, the \texttt{en\_core\_web\_trf} integrates a shared transformer layer that enhances contextual understanding and task consistency, significantly improving the precision of linguistic feature extraction. These advantages align closely with our system’s requirements, ensuring better detection and analysis of uncertainty. Detailed performance comparisons between available spaCy pipelines can be found in spaCy documentation\footnote{\href{https://spacy.io/models/en\#en\_core\_web\_trf}{https://spacy.io/models/en\#en\_core\_web\_trf}}. These benchmarks further validate the pipeline's suitability for our use case.

Additionally, the implementation of in-text citation standardization employs regular expressions (Regex). This process is crucial because there exist various variations of the in-text citation, such as "[1, 2, 5]", "(see Max \& Betty, 2002a; Marshal \& Mansell, 2001)", or "James et al. (2005)". During this process, the system converts all detected in-text citations to "@CITATION". Following this process, co-reference resolution is applied to maintain the narrative's continuity and coherence. This stage includes the identification and normalization of citation blocks, simplifying the task of distinguishing between the author(s)' arguments and those from former studies. The output now produces the clean text, ready for further processing.

The second step, Pattern Matching, employs a list of patterns derived from 12 SU pattern groups (see Table \ref{tab-examples}) with up to 89 patterns in total and a Scientific Uncertainty (SU) Lexicon that have been constructed during the pattern formulation process \ref{fig:patterns_formulation}. The input sentence is matched against the specified patterns. In the event that no match is found, the sentence is labeled 'Label 0' (Output 2), which means 'Claim'. Conversely, if a match is identified, a list of SU span candidates is generated, and the sentence is then moved to the next step. In order to optimize the matching process in this step, we have customized a spaCy rule-based matcher that takes into account keywords, span matches, sentence patterns, and linguistic features.

The third step, Complex Sentence Checking, determines whether any rebuttal, neutral, or confirmation statements are present in the sentence. The occurrences of these statements, along with the spans previously identified, eliminate the uncertainty in the sentence. If any of these patterns are detected, the sentence is labeled as 'Label 0' (Output 2), which is 'Claim', while if the system does not detect any such statements, the sentence is passed on to the final author reference checking step. Similar to the pattern matching step, this step uses a rule-based matcher from spaCy and patterns that capture the rebuttal or confirmation statement, as in the following examples:

\begin{quote}
\footnotesize
    $\textit{"...\textbf{no evidence to support} this hypothesis..."}$ \\\newline
    \textit{"\textbf{In order to test} whether X has a contribution to Y, \textbf{statistical analysis was employed}...."} \\\newline
    $\textit{"The high correlations scores \textbf{confirm} hypothesis H3"}$ \\
\end{quote}

Lastly, Authorial Reference Checking is conducted to determine the authorship of uncertainty expressions, distinguishing whether they originate from the current authors, prior studies, or a combination of both. In addition to utilizing the spaCy Matcher and SU Lexicon, Named Entity Recognition (NER) is used in this step to accurately identify direct mentions of author names from previous studies within the text. The outcome of this step is the annotation of sentences with authorial references. Once the process is complete, three outputs will be generated: sentence with 'Label 1' (Output 3), which is sentence with 'Uncertainty'; a list of final SU spans (Output 4), which is required to provide the interpretation of the annotation; and authorial reference label (Output 5). 

\ref{fig:appendix-collage} provides an overview of how UnScientify functions and its interface. The demonstration shows that out of the five sentences in the input text in the example, four express uncertainty. Two of these sentences discuss results or statements from previous studies. The detailed explanation provides information on which sentence elements support the outcome, as well as descriptive information on why the sentence is considered an SU expression. In the demonstration, the output identifies the first sentence as an SU expression due to the occurrence of the 'Explicit Scientific Uncertainty group' represented by the phrase 'remains unexplained' in the text. UnScientify also checks for authorial references and labels the first instance as 'Author(s)', indicating that the sentence originates from the author rather than being cited from other sources or previous studies. This provides more contextual and interpretable results. To test UnScientify, please visit the demo app at \url{https://bit.ly/unscientify-demo}.
       
\section{UnScientify Performance Evaluation}

In order to assess the performance of the UnScientify in detecting and analyzing uncertainty in scientific articles, we evaluated it on the corpora of 975 sentences described above. This section outlines the evaluation framework, as well as the comparative experimentation with Large Language Models (LLMs).

To contextualize the performance of the proposed system, comparative experiments were conducted with various LLMs: 1) Fine-tuning BERT-based language representation models, 2) Few-shot learning, and 3) Zero-shot scenario, using the same corpora of 975 annotated sentences. The Fine-tuning, Few-shot learning strategies, and Zero-shot scenario with prompting were designed to optimise the application of LLMs for this dataset and task. Table \ref{table:LLMs} shows the models used in the experiments.


\begin{table*}[htbp]
    \centering
    \footnotesize
   \caption{Models used in the experiments}
   \label{table:LLMs}
    \begin{tabularx}{\textwidth}{lXX}
        \toprule
        Experiment & Method & Model \\
        \midrule
        1 & \multirow[t]{100}{*}{\parbox{5.7cm}{\vspace{1.3\baselineskip}Fine-tuning BERT-based language representation models}} & a. SsciBERT \citep{shen_sscibert_2023} \\
        & & b. SciBERT \citep{beltagy_scibert_2019} \\
        & & c. DistilBERT\citep{sanh_distilbert_2019} \\
        & & d. RoBERTa \citep{liu_roberta_2019} \\
        2 & Few-shot learning & a. TARS \citep{halder_task-aware_2020} \\
        & & b. SetFit \citep{tunstall_efficient_2022} \\
        3 & Zero-shot learning & a. \multirow[t]{100}{*}{\parbox{8.5cm}{\vspace{1.1\baselineskip}comprehend-it-multilang-base \textsuperscript{*}}} \\
         & \\
        & & b. GPT4 \citep{openai_gpt-4_2023} \\
        & & c. Llama2  \citep{touvron_llama_2023} \\
        & & d. Mistral \citep{jiang_mistral_2023}  \\
        4 & weakly supervised with patterns matching & UnScientify \citep{eeke_2023}\\
        \bottomrule
    \end{tabularx}
\footnotesize 
\textsuperscript{*}\url{https://huggingface.co/knowledgator/comprehend_it-multilingual-t5-base}\\
\end{table*}


\addtocounter{table}{-1} 

\begin{table*}[h]
    \centering
    \footnotesize
    \caption{Distribution of datasets in each fold for Experiment 1}
    \label{table:data_experiment1}
    \begin{tabularx}{\textwidth}{XXXX}
        \toprule
        Dataset & Sentences with uncertainty & Sentences without uncertainty & Total number of sentences \\
        \midrule
        Training & $\sim$391 & $\sim$486 & $\sim$877 \\
        Validation & $\sim$43 & $\sim$55 & $\sim$98 \\
        \midrule
        Total & 434 & 541 & 975 \\ 
        \bottomrule
    \end{tabularx}
\end{table*}

\subsection{Fine-tuning BERT-based language representation models}

All 4 BERT-based models were fine-tuned using the Transformers library \citep{wolf_huggingfaces_2019}.
The BERT model generally showed high performance in text classification \citep{devlin_bert_2018}. At the same time, RoBERTa outperforms BERT in many benchmark tasks and DistilBERT is a distilled version of BERT retaining BERT's performance \citep{liu_roberta_2019, sanh_distilbert_2019}. SciBERT was pre-trained on a large corpus of scientific articles \citep{beltagy_scibert_2019} while SsciBERT was pre-trained on abstracts of scientific articles from the social sciences \citep{shen_sscibert_2023}.

The dataset used in this experiment consisted of 975 sentences, out of which 434 were labeled as containing uncertainty and 541 were labeled as containing non-uncertainty. To ensure a robust evaluation and a fair comparison among models, we implemented a stratified 10-fold cross-validation approach. This strategy preserved the class distribution within each fold, with 90~\% ($\sim$877 sentences) of the data allocated for training and 10~\% ($\sim$98 sentences) for validation. Table \ref{table:data_experiment1} shows the distribution of the data for the training and validation datasets. Each fold used identical training and validation splits across all models, ensuring consistency and comparability.

Training parameters were set as follows: learning rate = $5e^{-5}$, the evaluation is done (and logged) every evaluation step, the number of evaluation steps (number of update steps between two evaluations) is set to 40, the batch size per GPU for training and evaluation was set to 8, the total number of training epochs to perform was limited to 10.

\subsection{Few-shot learning}
Two approaches were used for the few-shot learning. Task-Aware Representation of Sentences (TARS) is a transformer-based approach for zero-shot or few-shot text classification \citep{halder_task-aware_2020}. TARS model considers semantic information of the class label, therefore we defined the labels in natural language. We used three sets of labels for this experiment, as Table~\ref{table:TARS} shows. We used the label "claim" for sentences that do not contain uncertainty, given that claim is one of the antonyms of uncertainty. 


\addtocounter{table}{-1} 

\begin{table*}[h]
    \centering
    \footnotesize
   \caption{Label sets used for the TARS fine-tuning}
   \label{table:TARS}
    \begin{tabularx}{\textwidth}{lXX}
        \toprule
        Set & Label 1 & Label 2 \\
        \midrule
        1 & uncertainty & claim \\
        2 & sentence containing uncertainty & sentence containing claim \\
        3 & Sentences containing uncertainty are sentences in which the author(s) express (explicitely) some lack of knowledge, or a lack of precision in the information, on a subject or a concept that is clearly identified & Sentences that do not contain uncertainty are sentences in which the authors do not express any uncertainty or lack of knowledge on the subject or concept that is treated. \\
        \bottomrule
    \end{tabularx}
\end{table*}

\clearpage

Sentence Transformers Fine-tuning (SetFit) is a framework designed to fine-tune sentence transformer models \citep{tunstall_efficient_2022}, enabling impressive performance with very limited training data. We used this framework to fine-tune a sentence transformer model \citep{reimers_sentence-bert_2019}, taking advantage of its ability to deliver effective results with as few as 8 labeled sentences. This makes SetFit particularly suitable for tasks with small datasets.

To improve the model's performance and capture a wider variety of linguistic patterns, we decided to use more training data than the minimum few-shot requirement. By splitting the dataset as shown in Table \ref{table:data_setfit}, we provided the model with a broader range of examples, ensuring it was exposed to different contexts and nuances of uncertainty.

\subsection{Zero-shot learning}

We tested different approaches for the zero-shot learning scenario on the whole dataset of 975 sentences. Google's mT5 model \footnote{\url{https://huggingface.co/knowledgator/comprehend_it-multilingual-t5-base}} \citep{xue_mt5_2021} serves as the foundation for the comprehend-it-multilang-base model, enabling its application in zero-shot learning. This model supports zero-shot learning through three different approaches: (1) employing entailment-based classification or Natural Language Inference (NLI) based classification with predefined labels, (2) performing text classification with  with Pre-defined Hypothesis, and (3) conducting NLI-based classification in Question-Answering Setup. In this study, the comprehend-it-multilang-base model was evaluated using these three approaches. First, it was tested with the same set of labels defined in natural language as in Experiment 2 (few-shot learning). Second, the model underwent text classification using an instruction tuning strategy, incorporating three prompts as shown in Table \ref{table:AI_questions} to provide additional context to form the instructions. Finally, the model was tested using the multiple choice question-answering approach, with the prompts used as questions.

\addtocounter{table}{-1} 

\begin{table*}[htbp]
    \centering
    \footnotesize
    \caption{Number of sentences in the training data in Experiment 2 (SetFit)}
    \label{table:data_setfit}
    \begin{tabularx}{\textwidth}{XXXX}
        \toprule
        Dataset & Sentences containing uncertainty & Sentences without uncertainty & Total number of sentences \\
        \midrule
        Train & 274 & 326 & 600 \\
        Test & 81 & 94 & 175 \\
        Validation & 79 & 121 & 200 \\
        Total & 434 & 541 & 975 \\ 
        \bottomrule
    \end{tabularx}
\end{table*}

In the second round of experiments, generative AI was examined: GPT4 \citet{openai_gpt-4_2023}, Llama2 \citet{touvron_llama_2023} and Mistral \cite{jiang_mistral_2023}. All models were tested with a set of three prompts, as Table~\ref{table:AI_questions} shows. Generally, prompts for conversational AI (specifically the GPT model) should contain instructions, context, input data and output indicator \cite{giray_prompt_2023, hu_improving_2024}. The prompts utilized in this study were built using these recommendations. Prompts 1 and 2 adhere to identical structures, but differ in wording, as we wanted to explore the potential impact of lexical choices on model accuracy. Prompt 3 is a condensed version of the previous prompts excluding the definition of uncertainty, which enables an examination of whether excessive instructions can affect the model's performance.  GPT4 was tested with two temperature settings: 0.2 and 1, while Llama2 and Mistral used the default temperature setting in Ollama's framework\footnote{\url{https://python.langchain.com/api_reference/ollama/chat_models/langchain_ollama.chat_models.ChatOllama.html}}.


\addtocounter{table}{-1} 

\begin{table*}[h]
    \centering
    \footnotesize
   \caption{Prompts used for the generative AI testing}
   \label{table:AI_questions}
    \begin{tabularx}{\textwidth}{lX}
        \toprule
        No. & Prompt \\
        \midrule
        1 & As an academic researcher, I would like to distinguish sentences that express scientific uncertainty in publications. Sentences containing uncertainty are sentences in which the author(s) express (explicitly) some lack of knowledge, or a lack of precision in the information, on a subject or a concept that is clearly identified. Does the following sentence contain scientific uncertainty? Respond in the form of yes or no. \\
        2 & As a scientist, I want to distinguish sentences that express scientific uncertainty in scientific articles. Sentences containing uncertainty are sentences in which the author(s) express (explicitly) some lack of knowledge, or a lack of precision in the information, on a subject or a concept that is clearly identified. Does the following text contain scientific uncertainty? Answer with yes or no. \\
        3 & As an academic researcher, I would like to distinguish sentences that express scientific uncertainty in publications. Does the following sentence contain scientific uncertainty? Respond in the form of yes or no. \\
        \bottomrule
    \end{tabularx}
\end{table*}

\section{Results}

\subsection{Models' performance comparison}
The experiments that we performed include three distinct experiments involving large language models (LLMs) and compare their performance with UnScientify. The results are detailed in Table \ref{table:LLMs evaluation}.

In the first experiment, RoBERTa emerged as the top-performing fine-tuned BERT-based model, achieving an accuracy of 0.799. This result highlights RoBERTa's superior ability to grasp the nuances of scientific language, along with its strong contextual understanding and adaptability.

The second experiment explored few-shot learning strategies, involving multiple configurations of Task-Aware Representation of Sentences (TARS) and Sentence Transformers Fine-tuning (SetFit) across 1 and 10 epochs. The analysis revealed that both SetFit configurations, regardless of the epoch count, attained the highest accuracy, each scoring 0.771. This result suggests a significant potential of the SetFit strategy in adapting to the task with limited labeled examples.

\addtocounter{table}{-1} 

\begin{landscape}
\footnotesize
\begin{longtable}{llllllll}
    \caption{Performances Evaluation} \label{table:LLMs evaluation} \\
    \toprule
    Experiment & Model & Label & Precision & Recall & F1-Score & Support & Accuracy \\
    \midrule
    \endfirsthead

    \multicolumn{8}{c}%
    {{\bfseries Table \thetable\ continued from previous page}} \\
    \toprule
    Experiment & Model & Label & Precision & Recall & F1-Score & Support & Accuracy \\
    \midrule
    \endhead

    \midrule
    \multicolumn{8}{r}{{Continued on next page}} \\
    \endfoot

    \bottomrule
    \endlastfoot

    1 & SciBERT & uncertainty & 0.641 & 0.6131 & 0.623 & 434 & 0.676\\
     & & claim & 0.705 & 0.726 & 0.714 & 541 & \\
     & SsciBERT & uncertainty & 0.789 & 0.765 & 0.771 & 434	& 0.798 \\
     &  & claim & 0.816 & 0.824 & 0.818 & 541 & \\
     & DistilBERT & uncertainty & 0.757 & 0.765 & 0.760 & 434	& 0.784 \\
     &  & claim & 0.810 & 0.799 & 0.802 & 541 & \\
     & RoBERTa & uncertainty & 0.762 & 0.811 & 0.780 & 434 &	\textbf{0.799} \\
     &  & claim & 0.847 & 0.789 & 0.813 & 541 & \\
     2 & TARS1 & uncertainty & 0.773 & 0.630	& 0.694 & 81 & 0.720 \\
     &  & claim & 0.755 & 0.819 & 0.786 & 94 & \\
     & TARS2 & uncertainty & 0.767 & 0.691 & 0.727 & 81 & 0.731 \\
     &  & claim & 0.763 & 0.787 & 0.775 & 94 & \\
     & TARS3 & uncertainty & 0.744 & 0.753 & 0.749 & 81 & 0.703 \\
     &  & claim & 0.772 & 0.755 & 0.763 & 94 & \\
     & SetFit (1 epoch) & uncertainty & 0.730 & 0.802 & 0.765 & 81 & \textbf{0.771} \\
     &   & claim & 0.814 & 0.745 & 0.778 & 94 & \\
     & SetFit (10 epoch) & uncertainty & 0.738 & 0.765 & 0.752 & 81 & 0.766 \\
     &   & claim & 0.791 & 0.766 & 0.778 & 94 \\
     3\textsuperscript{*} & GPT4 temp 0.2 (Prompt 1) & uncertainty & 0.690 & 0.724 & 0.706 & 434 & 0.732 \\
     &  & claim & 0.769 & 0.740 & 0.754 & 541 & \\
     & GPT4  temp 0.2 (Prompt 2) & uncertainty & 0.649 & 0.783 & 0.710 & 434 & 0.715 \\
     &  & claim & 0.792 & 0.660 & 0.720 & 541 & \\
     & GPT4  temp 0.2 (Prompt 3) & uncertainty & 0.647 &	0.850 & 0.735 & 434 & 0.727 \\
     &  & claim & 0.840 & 0.628 & 0.719 & 541 & \\
     & GPT4  temp 1 (Prompt 1) & uncertainty & 0.695 & 0.735 & 0.714 & 434 & \textbf{0.738} \\
     &  & claim & 0.777 & 0.741 & 0.759 & 541 & \\
     & GPT4  temp 1 (Prompt 2) & uncertainty & 0.649	& 0.793	& 0.714 & 434	& 0.717 \\
     &  & claim & 0.798 & 0.656 & 0.720 & 541 & \\
     & GPT4  temp 1 (Prompt 3) & uncertainty & 0.647 & 	0.850 & 0.735 & 434 & 0.727\\
     &  & claim & 0.840 & 0.628 & 0.719 & 541 & \\
     & knowledgator Approach1 label1 & uncertainty & 0.510 & 0.707 & 0.593 & 434 & 0.567\\
     &  & claim & 0.660 & 0.455 & 0.538 & 541 & \\
     & knowledgator Approach1 label2 & uncertainty & 0.503 & 0.553	& 0.527	& 434 & 0.558 \\
     &  & claim & 0.610 & 0.562 & 0.585 & 541 & \\
     & knowledgator Approach1 label3 & uncertainty & 0.498 & 0.276	& 0.356	& 434 & 0.554 \\
     &  & claim & 0.610 & 0.562 & 0.585 & 541 & \\
     & knowledgator Approach2 (Instruct-Prompt1) & uncertainty & 0.458 & 0.956 & 0.619 & 434 & 0.477\\
     &  & claim & 0.725 & 0.092 & 0.164 & 541 & \\
     & knowledgator Approach2 (Instruct-Prompt2) & uncertainty & 0.467 & 0.919	& 0.619	& 434 & 0.496 \\
     &  & claim & 0.708 & 0.157 & 0.257 & 541 & \\
     & knowledgator Approach2 (Instruct-Prompt3) & uncertainty & 0.483 & 0.371 & 0.420 & 434 & 0.544 \\
     &  & claim & 0.575 & 0.682 & 0.624 & 541 & \\
     & knowledgator Approach3 (QA-Prompt1) & uncertainty & 0.382 & 0.509 & 0.436 & 434 & 0.414\\
     &  & claim & 0.462 & 0.338 & 0.391 & 541 & \\
     & knowledgator Approach3 (QA-Prompt2) & uncertainty & 0.429 & 0.885	& 0.577	& 434 & 0.425 \\
     &  & claim & 0.375 & 0.055 & 0.097 & 541 & \\
     & knowledgator Approach3 (QA-Prompt3) & uncertainty & 0.331 & 0.249 & 0.284 & 434 & 0.442 \\
     &  & claim & 0.498 & 0.597 & 0.543 & 541 & \\
     & Llama2 7B Prompt 1 & uncertainty & 0.446 & 1.000 & 0.617 & 434 & 0.446 \\
     &  & claim & 1.000 & 0.001 & 0.003 & 541 & \\
     & Llama2 7B Prompt 2 & uncertainty & 0.449 & 0.998 & 0.619 & 434  & 0.454\\
     &  & claim & 0.909 & 0.018 & 0.036 & 541 & \\
     & Llama2 7B Prompt 3 & uncertainty & 0.446 & 1.000 & 0.617 & 434 & 0.448 \\
     &  & claim & 1.000 & 0.006 & 0.01 & 541 & \\
     & Mistral 7B Prompt 1 & uncertainty & 0.630	& 0.597 &	0.613 & 434 & 0.665 \\
     &  & claim & 0.690 & 0.719 & 0.704 & 541 & \\
     & Mistral 7B Prompt 2 & uncertainty & 0.660	& 0.518	& 0.581 & 434 & 0.667 \\
     &  & claim & 0.670 & 0.786 & 0.723 & 541 & \\
     & Mistral 7B Prompt 3 & uncertainty & 0.635 & 0.657 &	0.646 & 434 & 0.679 \\
     &  & claim & 0.717 & 0.697 & 0.707 & 541 & \\
    4 & UnScientify & uncertainty & 0.712 &	0.956 & 0.816 & 434 & \textbf{0.808} \\
     &  & claim & 0.952 & 0.689	& 0.800 & 541 & \\
\end{longtable}
\footnotesize 
    \vspace{2mm} 
    \begin{minipage}{\linewidth} 
        \textsuperscript{*}Note: Considering the probabilistic nature of the generative LLMs, which produce outcomes with multiple variations, each experiment setting in this group was executed three times for each model. The results presented in this table were derived from the majority of answers across the three runs.
    \end{minipage}
\end{landscape}

The third experiment assessed the performance of generative LLMs in a zero-shot setting by utilizing various prompts with GPT-4 at temperatures of 0.2 and 1, alongside LLaMA 2 7B, and Mistral 7B. While for the model from the Knowledgator, three different zero-shot approaches were conducted, including the experiment using a set of labels from experiment 2, performing text classification with instruction tuning using prompts, and conducting  question-answering approach using prompts. Considering the probabilistic nature of the generative LLMs which produce outcomes with multiple variations, each experiment setting in this group was executed three times for each model and setting. The results presented in Table \ref{table:LLMs evaluation} were derived from the majority of answers across the three runs.

The results from the third experiment show that GPT-4 with temperature settings of 1 (Prompt 1) achieved the highest accuracy of 0.738. This demonstrates the effectiveness of carefully crafted prompts and the influence of temperature settings on generative models' performance in zero-shot learning tasks. Meanwhile, the model from knowledgator also yields interesting results. As shown by Table \ref{table:LLMs evaluation}, the implementation of different zero-shot approaches to this model produced different results even though identical prompts (for Approach 2 and 3) were used during the process.

Next, UnScientify was evaluated using the full dataset. We processed all 975 sentences and compared the system's annotations to the ones in the dataset. The results show that our method for identifying uncertainty outperforms all other methods and the above-mentioned models, achieving an accuracy of 0.808. The system shows a very high precision for 'claim' labels (0.952) and recall for 'uncertainty' labels (0.956), which indicated that our approach is relevant for the processing of complex concepts such as scientific uncertainty.

Furthermore, we evaluated the performance of all tested models using the Friedman test. The normality of the data was examined using the Shapiro-Wilk test, and the homogeneity of variance was evaluated with Levene’s test. The p-values of both tests are below the 0.05 threshold, indicating that there exists a significant difference between the variances and that the data is not normally distributed. Consequently, we opted for the Friedman test over other alternative statistical methods such as ANOVA \citep{10.5555/1248547.1248548}. 

\addtocounter{table}{-1} 

\begin{table}[h]
\footnotesize
    \centering
    \begin{tabular}{ccc}
    \toprule
         Friedman chi-squared&  Degrees of Freedom & p-value\\
    \midrule
         1,160&  30& < 2.2e-16\\
    \bottomrule
    \end{tabular}
    \caption{Results of the Friedman test}
    \label{tab:friedman_res}
\end{table}

Zero-shot approaches were run three times each, therefore, the median prediction from the three outputs was used as the final prediction for evaluation. Additionally, fine-tuned models were evaluated using 10-fold cross-validation. Consequently, we combined the results across all folds. As Table~\ref{tab:friedman_res} demonstrates, the p-value of the Friedman test for classifiers is less than 0.05 indicating a statistically significant difference in the performance of the classifiers. 

\begin{figure*}[h]
    \centering
    \includegraphics[width=\textwidth]{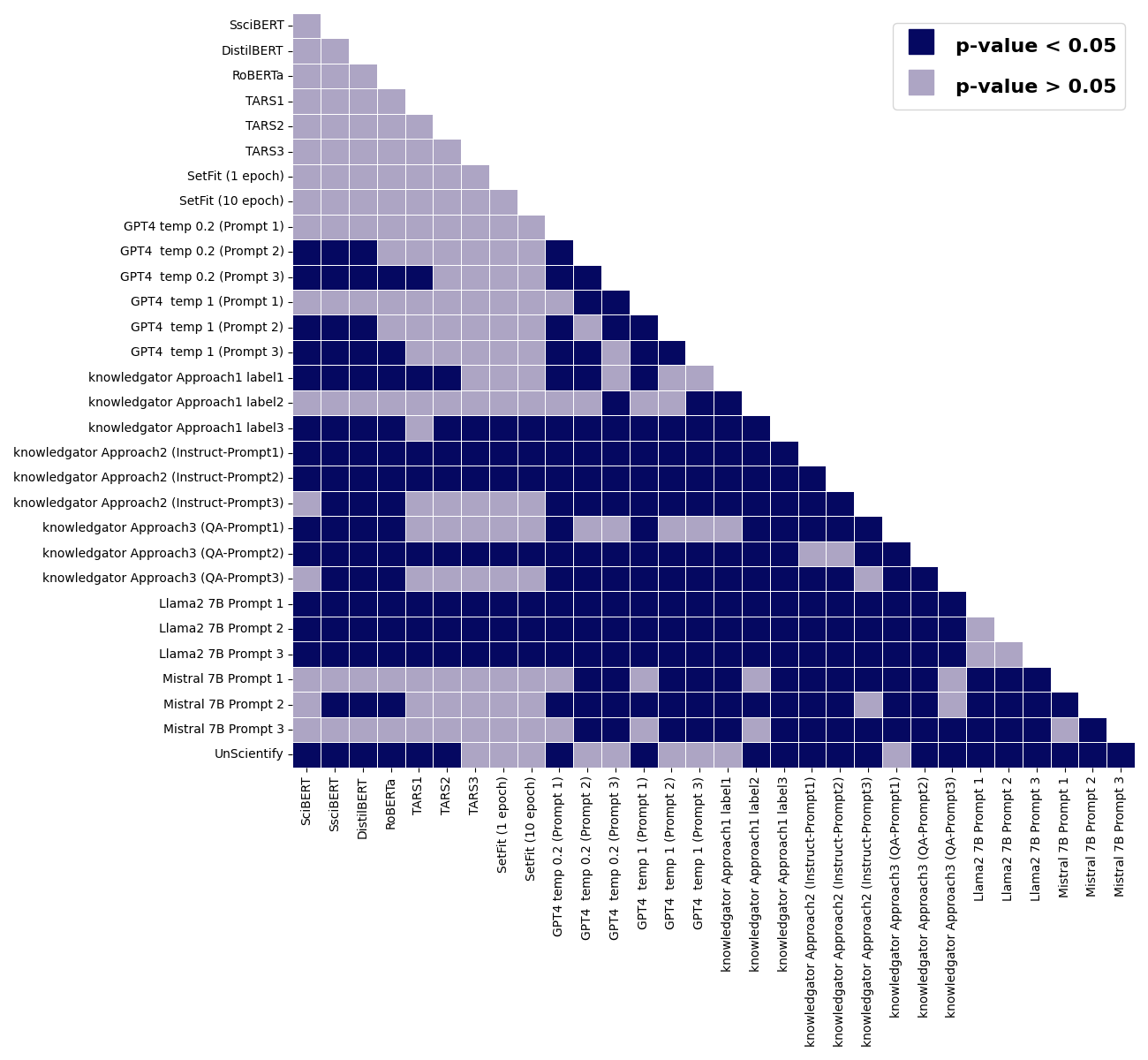}
    \caption{P-values of Pairwise Wilcoxon Rank Sum Test for classifier comparisons. P-values below the threshold of 0.05 indicate a statistically significant difference in the performance of classifiers. }
    \label{fig:Wilcoxon_visual}
\end{figure*}

To determine which classifiers exhibited statistically significant differences in performance, we conducted pairwise comparisons using the Pairwise Wilcoxon Rank Sum Test. To account for the increased risk of Type I error, the p-values were adjusted using the Bonferroni correction method for multiple comparisons. The test results are displayed in Figure~\ref{fig:Wilcoxon_visual}.

In the Pairwise Wilcoxon Rank Sum Test, the null hypothesis posits that there is no significant difference between the two groups, with a significance level set at 0.05. As Table~\ref{tab:Wilcoxon_UnScientify} shows there is a statistically significant difference between the performance of UnScientify and all classifiers except TARS3, SetFit, GPT 4 prompts 2 and 3, knowledgator Approach1 label1, and knowledgator Approach3 (QA-Prompt1).

\begin{table}[h]
    \centering
    \footnotesize
    \caption{P-values of Pairwise Wilcoxon Rank Sum Test for the comparison of UnScientify and other models.}
    \label{tab:Wilcoxon_UnScientify}
\begin{tabular}{lr}
\toprule
                                   Model &      p-value \\
\midrule
                                  SciBERT & \textbf{1.777101e-14} \\
                                 SsciBERT &\textbf{ 1.804134e-19} \\
                               DistilBERT &\textbf{ 1.947215e-15} \\
                                  RoBERTa & \textbf{7.117186e-11} \\
                                    TARS1 & \textbf{2.850555e-04} \\
                                    TARS2 & \textbf{0.016  }\\
                                    TARS3 & 0.748  \\
                         SetFit (1 epoch) & 1 \\
                        SetFit (10 epoch) & 1 \\
                 GPT4 temp 0.2 (Prompt 1) & \textbf{4.701133e-11} \\
                GPT4  temp 0.2 (Prompt 2) & 0.253  \\
                GPT4  temp 0.2 (Prompt 3) & 1 \\
                  GPT4  temp 1 (Prompt 1) & \textbf{1.851902e-10} \\
                  GPT4  temp 1 (Prompt 2) & 0.979 \\
                  GPT4  temp 1 (Prompt 3) & 1 \\
            knowledgator Approach1 label1 & 1 \\
            knowledgator Approach1 label2 & \textbf{2.998742e-04} \\
            knowledgator Approach1 label3 & \textbf{1.579781e-49} \\
knowledgator Approach2 (Instruct-Prompt1) & \textbf{2.605883e-57} \\
knowledgator Approach2 (Instruct-Prompt2) & \textbf{2.802330e-40} \\
knowledgator Approach2 (Instruct-Prompt3) & \textbf{5.656876e-28} \\
      knowledgator Approach3 (QA-Prompt1) & 1 \\
      knowledgator Approach3 (QA-Prompt2) & \textbf{8.182360e-48} \\
      knowledgator Approach3 (QA-Prompt3) & \textbf{5.845618e-24} \\
                       Llama2 7B Prompt 1 &\textbf{ 2.343425e-84} \\
                       Llama2 7B Prompt 2 &\textbf{ 2.563024e-81} \\
                       Llama2 7B Prompt 3 &\textbf{ 1.731503e-83} \\
                      Mistral 7B Prompt 1 &\textbf{ 1.166394e-15} \\
                      Mistral 7B Prompt 2 &\textbf{ 5.954896e-30} \\
                      Mistral 7B Prompt 3 & \textbf{8.774242e-10} \\
\bottomrule
\end{tabular}
\end{table}

\subsection{Error Analysis}

Error analysis is a crucial component in evaluating the performance and robustness of language models. In this section, we conduct an error analysis to assess the accuracy and reliability of the model's predictions.

The implementation of LLMs for text annotation, particularly in identifying scientific uncertainty from scientific articles, presents several challenges and inconsistencies. A significant concern is the inherent "black-box" nature of LLMs, which makes it difficult to trust the outputs produced by these models. While fine-tuned LLMs such as SciBERT appear promising as tools for annotating scientific uncertainty, the rationale behind its predictions remains opaque. 

\begin{figure*}[h]
    \centering
    \includegraphics[width=1.1\linewidth]{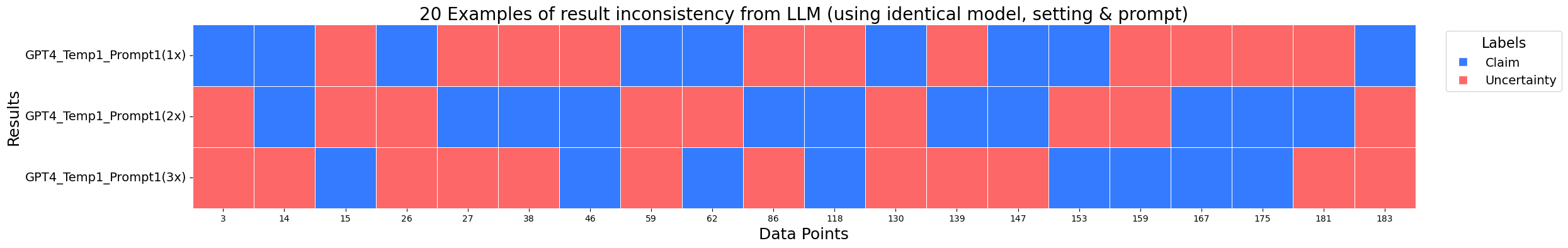}
    \caption{Examples of result inconsistency from LLM (using identical model, setting \& prompt)}
    \label{fig:GPT4_inconsistency}
\end{figure*}

Generative LLMs such as GPT-4, the model from Knowledgator, Llama2, and Mistral also exhibit questionable results regarding the reasoning behind their predictions. When we re-ran the experiments three times for each model in this group, the results indicated inconsistency. Table \ref{table:LLMs_inconsistency} details the occurrences of these inconsistencies across three runs for each model in each experimental setting. Furthermore, Figure \ref{fig:GPT4_inconsistency} illustrates some examples of inconsistent results from the best performer in this experiment group: GPT-4 with temperature 1 using Prompt 1. 

\clearpage
\begin{table*}[h]
    \centering
    \footnotesize
    \caption{Inconsistencies in LLM labeling following three labeling processes}
    \label{table:LLMs_inconsistency}
    \begin{tabularx}{\textwidth}{XX}
        \toprule
        Model & Inconsistency (n) \\
        \midrule
        GPT4 temp 1 (Prompt 1) & 98 \\
        GPT4 temp 1 (Prompt 2) & 113 \\
        GPT4 temp 1 (Prompt 3) & 82 \\
        GPT4 temp 0.2 (Prompt 1) & 49 \\
        GPT4 temp 0.2 (Prompt 2) & 65 \\ 
        GPT4 temp 0.2 (Prompt 3) & 35 \\ 
        knowledgator (All Approaches) & 0 \\ 
        Llama2 7B Prompt 1 & 2 \\ 
        Llama2 7B Prompt 2 & 6 \\ 
        Llama2 7B Prompt 3 & 19 \\ 
        Mistral 7B Prompt 1 & 83 \\ 
        Mistral 7B Prompt 2 & 54 \\ 
        Mistral 7B Prompt 3 & 60 \\ 
        \bottomrule
    \end{tabularx}
\end{table*}

As illustrated in Figure \ref{fig:GPT4_inconsistency}, GPT-4 model at temperature 1 and Prompt 1 resulted in inconsistent results. We use the text from Data point 14 as an example:

\begin{quote}
    \textit{"This motivates a new hypothesis, that sensory memories can act offline (indirectly) on sensorimotor performance via spontaneous activity."}
\end{quote}

In the first and second runs, GPT4 annotated this text as 'claim', and in the third run, it labeled the text as 'uncertainty'. These results make it difficult for us to identify the correct label and the cause of error in this model. This behavior also occurred in other generative LLMs, except for the model from knowledgator.

\begin{figure*}[htbp]
    \centering
    \includegraphics[width=1.1\linewidth]{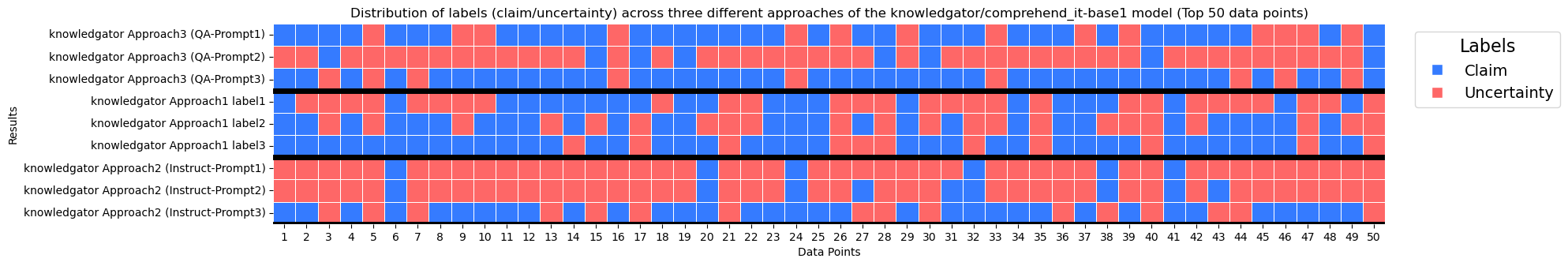}
    \caption{Distribution of labels (claim/uncertainty) across three different zero-shot approaches of the knowledgator/comprehend\_it-base1 model (top 50 data points)}
    \label{fig:knowledgator_inconsistency}
\end{figure*}

Nevertheless, the model from knowledgator revealed another kind of questionable results. The model showed inconsistent results when different zero-shot approaches were implemented as depicted by Figure \ref{fig:knowledgator_inconsistency}. For example, identical prompts used in Approach 2 (zero-shot instruction) and Approach 3 (zero-shot question-answering) yielded different predictions for some data points. Moreover, the results from those two approaches showed different results to the results from approach 1 (zero-shot NLI-based classification with predefined labels) as shown in Figure \ref{fig:knowledgator_inconsistency}. The inconsistent labels (shown by the colors) happen in the same data point because of the different zero-shot approaches used.

Conversely, the UnScientify did not exhibit any inconsistency issues. Its predictions remained stable across multiple runs. However, errors still emerged, primarily due to the absence of certain scientific uncertainty expression patterns in the SU Patterns Groups. Below is the sample cases that UnScientify inaccurately annotated.

\begin{quote}
    \textit{"The study \textbf{needs to be replicated} in different settings using a larger sample size \textbf{to ensure generalizability}."}
\end{quote}
\begin{quote}
    \textit{"Nonetheless, \textbf{only a subset of} alcohol consumers develops CRC."}
\end{quote}

These errors were typically associated with uncommon indirect expressions in the text which the model had never seen previously. To deal with these errors, we only need to add new patterns to the SU Patterns Group or adjust the available patterns to make UnScientify able to accommodate these writings and capture these expressions.

\section{Discussion}
\subsection{Comparative Insights from Alternative Approaches}
The comparative analysis of the results underscores the distinct challenges and advantages of each approach in the context of uncertainty detection in scientific texts. The inclination towards employing LLMs for tasks such as uncertainty detection is understandable, given their remarkable success in various NLP tasks. However, the deployment of LLMs comes with its own set of challenges \citep{sarkar2024llms}, including substantial computational requirements and the need for significant customization to adapt to specific tasks like uncertainty detection.

The statistical analysis of the results revealed a significant difference in performance among the tested approaches. However, no statistically significant differences were observed between the fine-tuned models, which indicates that observed differences may occurred by chance. Notably, the performance of the fine-tuned models was significantly different to that of most zero-shot approaches, including generative models such as GPT, Llama, and Mistral.


The fine-tuning scenario with BERT-based models highlights that while domain-specific pre-training can be advantageous, its effectiveness depends on how closely the task aligns with the nuances of the pre-training corpus. Among the four models, RoBERTa achieved the highest accuracy (0.799), followed closely by SsciBERT (0.798), while DistilBERT (0.784) and SciBERT (0.675) lagged behind. Despite both SciBERT and SsciBERT being pre-trained on scientific texts, SsciBERT's strong performance suggests its social science pre-training effectively captured patterns relevant to uncertainty detection. In contrast, SciBERT may have overfit to its pre-training data, limiting its adaptability to the task.

RoBERTa’s superior performance stems from its advanced pre-training strategies, including larger datasets in the pre-training process, dynamic masking, and removal of Next Sentence Prediction (NSP), which enhance its ability to handle nuanced linguistic tasks. DistilBERT, while efficient and lightweight, showed weaker performance, likely due to its reduced model complexity, which limits its capacity to capture subtle cues. These results reveal that general-purpose models like RoBERTa excel in tasks requiring complex linguistic reasoning, while domain-specific models like SsciBERT perform well when their pre-training aligns closely with the task, and lighter models like DistilBERT prioritize efficiency over depth.

Few-shot learning strategies, particularly SetFit, demonstrated promising adaptability with limited data, suggesting a potential pathway for tasks constrained by sparse labeled examples. Nevertheless, the dependency on the quality and representativeness of few-shot examples could limit the approach's robustness across broader applications. These methods still face the challenge of transferability, where the model's ability to generalize from a few examples to unseen data can vary significantly based on the task complexity and the specificity of the examples provided. The performance of few-shot learning significantly depends on the choice of examples \citep{NEURIPS2021_5c049256}. Poorly selected examples can lead to suboptimal model understanding and performance. Additionally, crafting effective prompts can require substantial domain expertise and manual effort. Therefore, employing these strategies requires a balance between leveraging pre-trained knowledge and adapting to the specifics of the task at hand, often necessitating fine-tuning and experimentation. The weakly supervised approach that we propose, in contrast, leverages existing patterns and rules that can be easily adjusted or expanded upon by domain experts without the need for extensive datasets or complex model tuning.

The zero-shot learning scenario revealed the power of generative LLMs to adapt to tasks without direct training, albeit with the caveat of prompt sensitivity and the need for temperature adjustments to optimize performance. While promising, the variability in outcomes based on prompt design and model configurations underscores the importance of experimentation to identify optimal setups. Moreover, the interpretability of the decisions made by generative LLMs in a zero-shot context remains a challenge, as the reasoning behind their outputs can be opaque because of the "black box" nature of these models \citep{singh2024rethinking}, making it difficult to trust and verify their assessments in critical applications such as scientific research. This lack of transparency can be a significant drawback in scientific and technical applications where understanding the "why" behind a prediction is as important as the prediction itself. In the case of UnScientify, the weakly supervised method, with its transparent and interpretable framework, offers clarity and ease of adjustment that generative LLMs currently struggle to match.

The Wilcoxon Rank Sum Test further revealed statistically significant differences between UnScientify's performance and that of most classifiers, except for certain zero-shot approaches such as TARS3, SetFit, GPT-4 prompts 2 and 3, and specific configurations of Knowledgator (Approach1 label1 and Approach3 QA-Prompt1). In cases where no significant difference was observed, this outcome may be partially attributed to the limited size of the testing dataset, which can reduce statistical power and obscure meaningful performance distinctions. Despite this, UnScientify demonstrates a distinct advantage in its ability to efficiently handle limited labeled datasets through its rule-based, weakly supervised approach. Unlike data-intensive machine learning or zero-shot LLM methods, UnScientify's reliance on interpretable linguistic patterns and minimal annotated data allows for quick adaptation and effective uncertainty detection without extensive retraining or resource demands.

\subsection{Performance and Insights from the UnScientify System}


UnScientify, the weakly supervised system introduced in this study, exemplifies a balanced and effective approach to detecting uncertainty in scientific texts. By combining a rule-based pattern matching strategy with a mechanism for complex sentence analysis, the system achieves an accuracy of 0.808,  with a good balance between precision, recall, and F1 scores. This performance underscores UnScientify’s ability to navigate the complex linguistic structures often found in scientific discourse. At the same time, this approach is lightweight, and it does not require large labeled datasets or intensive computational resources.

\subsubsection{UnScientify vs. Transformer-Based Language Models}


Transformer-based language models excel in semantic and contextual understanding by leveraging token embeddings (derived from word or subword units) and, in some cases, sentence embeddings to represent entire sentences as single vectors. These embedding strategies have proven highly effective in numerous NLP tasks, offering state-of-the-art performance in areas like text classification, named entity recognition, and semantic similarity. However, these approaches face significant challenges in tasks like phrase detection, similar to our study, which heavily depend on accurately identifying uncertainty assertion spans within sentences.

Word or token embeddings, as utilized in Transformer models, benefit from semantic representation and attention mechanisms, allowing them to capture relationships between tokens and spans effectively. However, these embeddings are still limited in explicitly isolating and representing critical spans within sentences \citep{joshi-etal-2020-spanbert}, particularly in tasks that require a precise focus on specific spans amid noisy or domain-specific contexts. While the attention mechanism aids in contextualizing each token, the resulting embeddings may diffuse focus across the sentence, making it challenging to zero in on key spans essential for certain applications, such as identifying uncertainty expressions. On the other hand, sentence embeddings, like those used in SetFit models, condense an entire sentence into a single dense vector. This approach is highly effective for tasks that benefit from a holistic sentence-level understanding, but it poses significant limitations for applications that require precise retrieval of sub-sentence phrases. By encoding the entire sentence as a single unit and representing a complete sentence with a single dense vector, sentence embeddings risk losing the nuanced relationships and interdependencies within smaller spans, making them less suitable for span-specific tasks. This limitation has been highlighted by \cite{orbach2024spanaggregatablecontextualizedwordembeddings}, who observed that in real-world phrase retrieval applications, dense sentence representations struggle when target phrases reside within noisy contexts. Similarly, in our study, these challenges impede the ability of transformer-based models to precisely focus on uncertainty assertion spans, particularly when surrounded by domain-specific or extraneous information, and when the training data are limited. For example: 



\begin{quote}
     \textit{"The correlation is often unclear between recovery speed from Covid-19 and diet consumption"}
\end{quote}

The sentence embedding-based representations might tend to dilute the signal in the sentence. Text classifiers can overemphasize common, non-discriminatory phrases such as \textit{"recovery speed from Covid-19 and diet consumption"} obscuring the core uncertainty span \textit{"The correlation is often unclear"} particularly in complex or verbose sentences.

Our UnScientify system, in contrast, focuses on linguistic patterns and features, which allows us to detect uncertainty expressions consistently regardless of the surrounding content or domain. By operating independently of domain-specific training, UnScientify is more efficient in capturing the important spans of scientific uncertainty expression from texts, providing a practical solution for uncertainty detection without the computational and data demands of training or fine-tuning large models.

Another key strength of our approach lies in its ability to provide interpretable outputs. Unlike black-box models, UnScientify can provide the reasons why specific sentences are flagged as uncertain, identifying and highlighting the linguistic elements from the evaluated sentences that contribute to this categorization. This transparency enhances trust in the system's outputs and makes it a valuable tool for researchers who seek not only results but also an understanding of the reasoning behind them.


\subsubsection{UnScientify vs. Former Rule-Based Systems}


UnScientify’s integration of linguistic features makes it more flexible and effective than former rule-based systems that mostly utilized token level identification \citep{omero_writers_2020} or regex-based system \citep{chapman_simple_2001, HARKEMA2009839, atanassova_studying_2018}. Token-level detection approaches struggle to accurately identify scientific uncertainty expressions because they fail to capture broader syntactic and semantic relationships within a sentence. Similarly, regex-based systems rely solely on character-based pattern matching, lacking the ability to parse grammatical dependencies, which further limits their effectiveness in detecting linguistic nuances.

In contrast, UnScientify is capable of processing complex sentence structures involving dependency relationships. This includes the identification of uncertainty expressions in embedded clauses or across non-contiguous spans. As was found by \cite{peng2017negbiohighperformancetoolnegation}, using dependency to detect negation or uncertainty expression in radiology reports made the detection 9.5~\% higher in precision and 5.1~\% higher in F1-score compared to using a regex-based approach.





Another feature of UnScientify is its ability to annotate authorial references, distinguishing whether expressions of uncertainty stem from the current author(s), previous studies, or a combination of sources. While this functionality has not yet been formally evaluated, it represents a significant step forward in uncertainty analysis, with the possibility of  examining the consensus or the divergences within specific research domains.


\section{Limitations}

While UnScientify presents a viable and effective solution for detecting and analyzing uncertainty in scientific articles, several limitations should be noted. The system relies on pattern matching, which, although effective, may not fully capture the complexity of all linguistic expressions of uncertainty. However, its design offers a significant advantage: UnScientify is very easy to maintain and improve. Unlike machine learning models or LLMs that require computationally intensive retraining, enhancing UnScientify’s performance only involves adding new patterns to its system. This flexibility allows it to adapt quickly to new requirements or domains without the need for large datasets or retraining.

Furthermore, while UnScientify is inherently domain-agnostic and leverages linguistic patterns, increasing the size and diversity of the dataset could introduce valuable variations to enrich its pattern library. The current dataset, which covers Medicine, Biochemistry, Interdisciplinary and Empirical Social Sciences, provides a strong foundation as it represents Science, Technology and Medicine (STM), Social Sciences and Humanities (SSH), and two journals (Nature and PLoS One) representing both broad and interdisciplinary research. However, we recognize that an extension to other fields such as engineering or arts and humanities could further improve the generalization of the system.

Although optimized for English scientific texts, Unscientify's rule-based approach is not easily transferable to other languages or tasks without significant adaptation. Despite these limitations, its evaluation scores highlight its superior capability to identify uncertainty expressions compared to other models, while also providing interpretable results. This makes UnScientify a practical, efficient, and reliable tool for uncertainty detection, offering an easy-to-maintain alternative to computationally intensive LLMs. Additionally, while GPT-4 was included in comparisons, the reproducibility of such experiments is not guaranteed due to potential modifications to GPT-4 or discontinuation of access to specific versions.


\section{Conclusion}
\subsection{Key Findings}
UnScientify offers a first comprehensive approach to identifying uncertainty expressions in scientific text. By utilizing pattern matching, complex sentence checking, and authorial reference checking, it provides clear and interpretable output that explains why a sentence is annotated as expressing uncertainty, addresses the element of SU expression, and verifies authorship reference.

The results from the evaluation show that UnScientify outperforms all other tested models for the task of identifying scientific uncertainty in sentences. These results show that our method presents a potential for enhancing information retrieval, text mining, and broader scientific article processing. Moreover, it lays the groundwork for further research on scientific uncertainty. To further enhance the UnScientify system, we intend to extend the annotation to more dimensions of scientific uncertainty including its nature, context, timeline, and communication characteristics. In future work, we also aim to explore hybrid approaches that combine spaCy’s linguistic analysis with the semantic depth of large language models (LLMs), which may further enhance the system's performance and adaptability.

\subsection{Contributions}
The key contributions of this study are as follows:
\begin{itemize}
    \item \textbf{UnScientify System}: A weakly supervised system combining linguistic patterns, contextual analysis, and interpretable outputs for detecting scientific uncertainty.
    \item \textbf{Gold-Standard Dataset}: AURORA-MESS\footnote{\url{https://zenodo.org/records/15001250}}, a publicly available annotated corpus to support reproducibility and further research.
\end{itemize}

\begin{acknowledgment}
This research was funded by the French ANR InSciM Project (2021-2025)\footnote{\href{https://project-inscim.github.io/}{https://project-inscim.github.io/}} under grant number ANR-21-CE38-0003-01, and the Chrysalide Mobilité Internationale des Doctorants (MID) mobility grant from the University of Bourgogne Franche-Comté, France. Additionally, our appreciation extends to the GESIS -- Leibniz Institute for the Social Sciences for providing the dataset and invaluable support.
Nina Smirnova received funding from the Deutsche Forschungsgemeinschaft (DFG) under grant number: MA 3964/7-3 (POLLUX project).
Philipp Mayr received additional funding by the European Union under the Horizon Europe grant OMINO – Overcoming Multilevel INformation Overload\footnote{\href{http://ominoproject.eu}{http://ominoproject.eu}} under grant number 101086321 \citep{holyst024}.
\end{acknowledgment}

\bibliographystyle{plainnat} 
\bibliography{references} 

\clearpage 

\appendix

\begin{center}
\LARGE \textbf{Appendices}
\end{center}



\section*{Appendix A}
\stepcounter{section} 
\appendixnumbering
\centering
\footnotesize
\begin{longtable}{p{2.5cm}p{6cm}p{7.5cm}}
\caption{SU Pattern Groups and examples of annotated sentences with SU spans written in bold}\label{tab-examples} \\ 
\toprule
\textbf{Pattern Group} & \textbf{Description} & \textbf{Examples} \\
\midrule
\endfirsthead
\toprule
\textbf{Pattern Group} & \textbf{Description} & \textbf{Examples} \\
\midrule
\endhead
\midrule
\multicolumn{3}{r}{{Continued on next page}} \\
\endfoot
\bottomrule
\endlastfoot

Explicit SU & Explicit SU group displays expressions with obvious scientific uncertainty keywords, indicating direct and explicit uncertainty expression & 1) In addition, the role of the public \textbf{is often unclear}. \newline 2) ... the functional relevance of G4 in vivo in mammalian cells \textbf{remains controversial}. \\

Modality & The modality group comprises expressions that indicate uncertainty through the use of modal language & 1) Different voters \textbf{might have} different interpretations about ... \newline 2) There \textbf{may also be} behavioral effects. \\

Conditional Expression & The conditional expression group includes expressions that indicate uncertainty by presenting a condition or circumstance that must be met for a certain outcome to occur & 1) \textbf{If} persons perceive the media as hostile, \textbf{it is probable that} the mere-exposure effect is weakened thus we hypothesize... \newline 2) \textbf{If} there are any violations, subsequent inferential procedures may be invalid, and \textbf{if so}, the conclusions would be faulty. \\

Hypothesis & The hypothesis group encompasses expressions that indicate uncertainty by proposing a tentative explanation or assumption that requires further testing and verification to be confirmed & 1) \textbf{Hypotheses predict that} aggregate support for markets should be stronger... \newline 2) \textbf{We assume} that post-materialistic individuals may have differing attitudes towards doctors than those... \\

Prediction & The prediction group comprises expressions that indicate uncertainty by proposing a forecast or projection that may or may not come to fruition, thereby introducing an element of uncertainty & 1) In July 2017, the National Grid's Future Energy Scenarios \textbf{projected that} the UK government... \newline 2) Since aging leads to decreased Sir2, we \textbf{predicted that}, in young cells... \\

Interrogative Expression & The interrogative expression group includes expressions that indicate uncertainty by posing a question or series of questions, which may suggest doubt or uncertainty about a particular concept or phenomenon & 1) The study aims to determine \textbf{whether} the observed results can be replicated across different populations. \newline 2) ...this research literature has also contested \textbf{whether or not} citizens' knowledge about these issues is accurate... \\

Non-generalizable statement & The non-generalizable statement group expresses uncertainty with limited scope or applicability, which may not represent a broader context or population & 1) Our study ... thus \textbf{cannot be directly generalized} to low-income nations nor extrapolated into the long-term future. \newline 2) ...estimates \textbf{may not be generalisable} to women in other ancestry groups... \\

Adverbial SU & The scientific uncertainty group includes adverbs that modify or shift the sentence's meaning, introducing uncertainty & 1) ...direct and indirect readout during the transition from search to recognition mode is \textbf{poorly} understood. \newline 2) It will be \textbf{quite} certain that they belong to the subpopulation of gender heterogenous... \\

Negation & The negation group comprises expressions that indicate uncertainty through the use of negation which may alter the meaning of the sentence and introduce an element of uncertainty & 1) The identity of C34 modification in... is \textbf{not clear}. \newline 2) There was \textbf{no consistent} evidence for a causal relationship between age at menarche and lifetime number of sexual partners... \\

Subjectivity & The subjectivity group includes expressions indicating uncertainty through subjective language like opinions, beliefs, or personal experiences & 1) \textbf{We believe that} there are good reasons for voters to care about... \newline 2) \textbf{To our knowledge}, this is the first study to provide global... \\

Conjectural & The conjectural group expresses uncertainty through conjecture or speculation, using guessing or suppositions without concrete evidence & 1) This belief \textbf{seems to be} typical for moderate religiosity. \newline 2) Better performance \textbf{seems to be linked} to life satisfaction, suggesting that individuals... \\

Disagreement & The disagreement group includes expressions that express uncertainty through disagreement or contradiction, often indicating opposing viewpoints or conflicting evidence & 1) \textbf{In contrast to previous studies}, our results did not show a significant effect... \newline 2) \textbf{On the one hand}, some researchers argue that the use of technology in the classroom can enhance... \\

\end{longtable}

\FloatBarrier 

\section*{Appendix B}
\stepcounter{section} 
\appendixnumbering
\begin{figure}[H]
\centering
\begin{minipage}{.5\textwidth}
  \centering
  \includegraphics[width=1\linewidth]{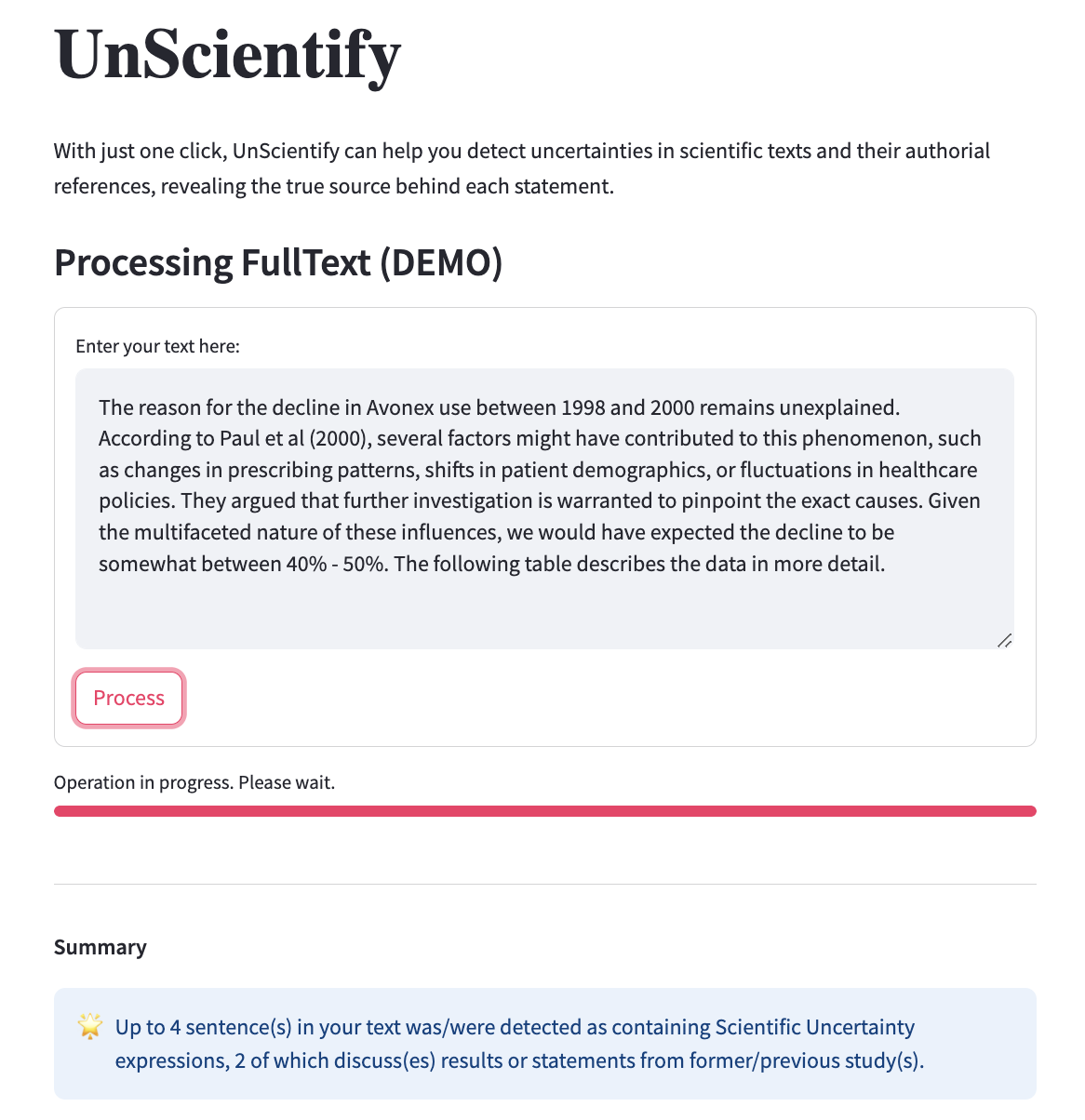}
  \label{fig:app1}
\end{minipage}%
\begin{minipage}{.5\textwidth}
  \centering
  \includegraphics[width=1.1\linewidth]{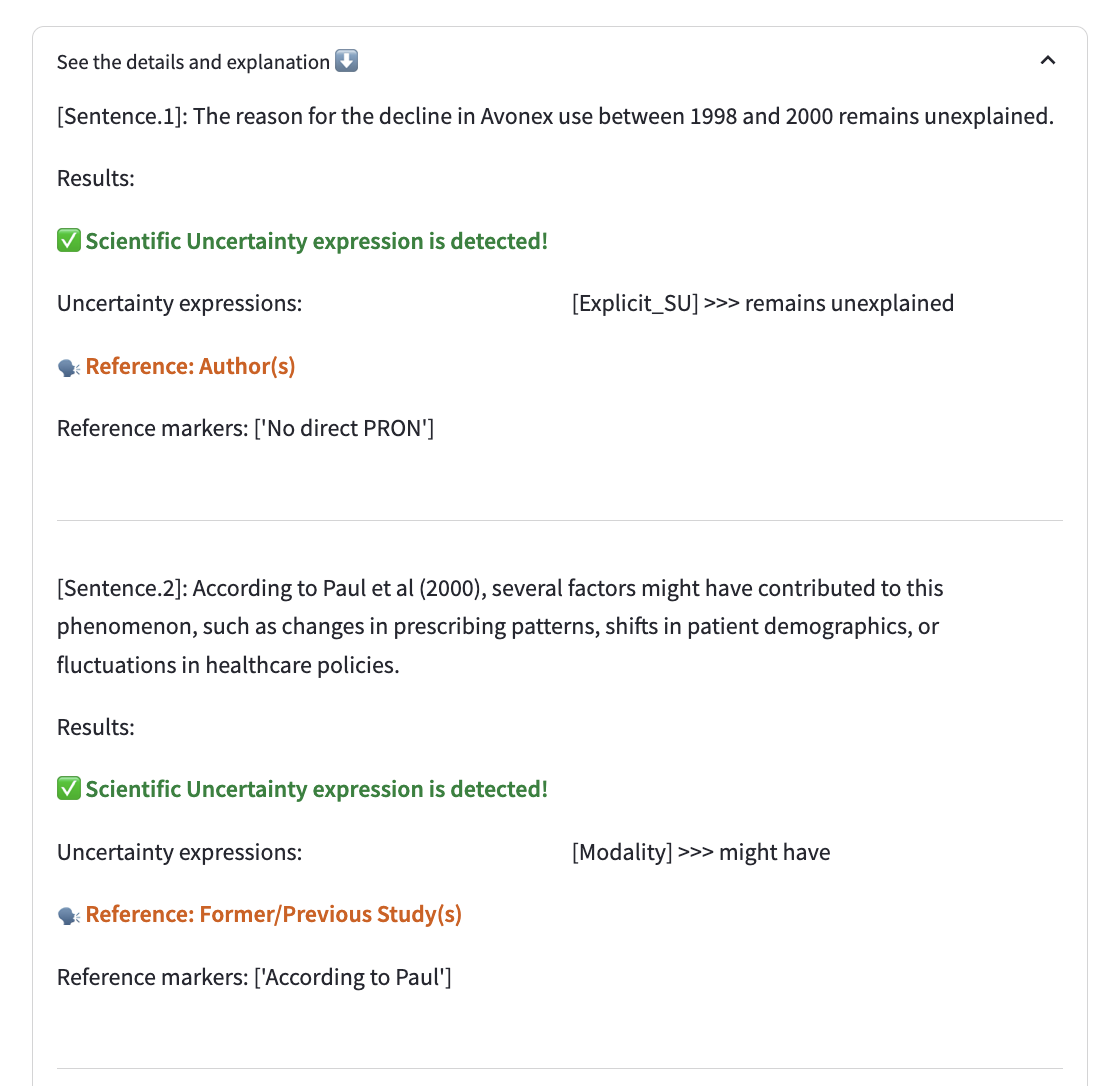}
  \label{fig:app2}
\end{minipage}

\begin{minipage}{.5\textwidth}
  \centering
  \includegraphics[width=1\linewidth]{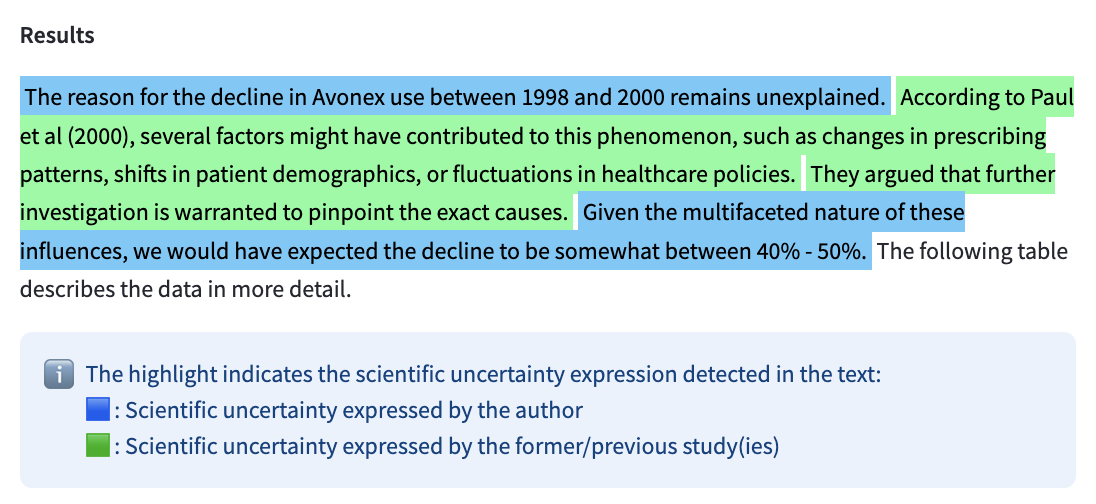}
  \label{fig:app3}
\end{minipage}%
\begin{minipage}{.5\textwidth}
  \centering
  \includegraphics[width=1.1\linewidth]{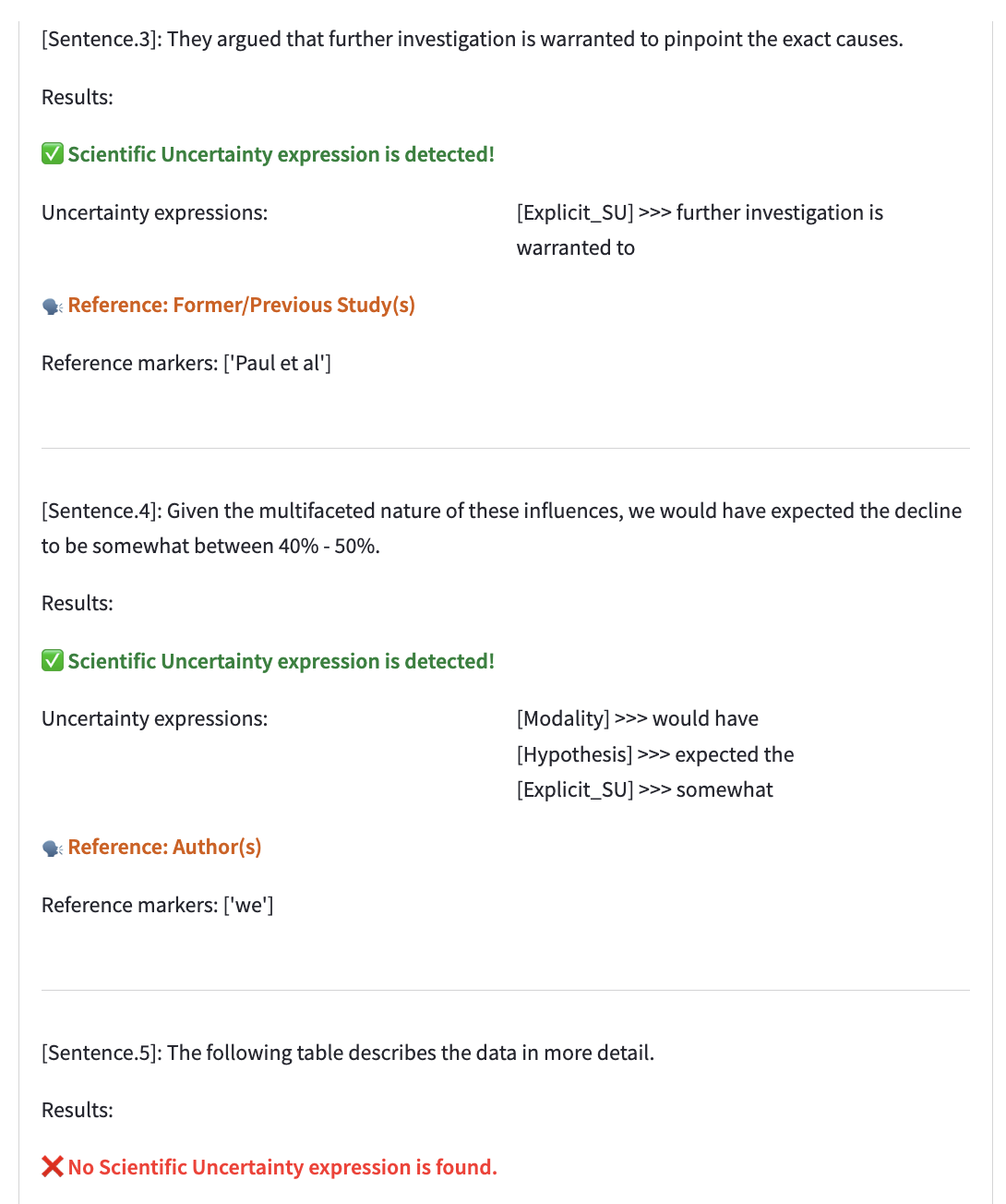}
  \label{fig:app4}
\end{minipage}
\caption{UnScientify Interface and Output}
\label{fig:appendix-collage}
\end{figure}

\FloatBarrier 

\section*{Appendix C}

\stepcounter{section} 

\begin{table}[H]
\footnotesize
\centering
\caption{The list of cues that compose the data for data sampling with the cue mapping method}
\label{tab:Cues_list}
\begin{tabular}{lp{3cm}p{8.5cm}}
\toprule
Source & Category & Cues / Markers \\
\midrule
\citet{hyland_talking_1996} & – & Would (not); may (not); could; might (not); should; cannot; will (not); must; shall; ought to \\
\hline
\citet{chen_scalable_2017} & – & Unclear; suspect; controversial; ambiguity; inconclusive; unexpected; consensus; contrary; inconsistent; paradoxical; confusing; unusual; uncertain; flaw; uncertainty; dispute; unknown; impossible; ambiguous; misleading; incomplete; unexplained; contradictory; contentious; paradox; incompatible; surprising \\
\hline
\multirow{3}{*}{\citet{bongelli_writers_2019}} & Epistemic verbs & I suppose; suggest/s; seem/s; suggesting; assuming; we think; I believe; it seems; expect; appear; look/s; suspect/suspected; do not seem; no one has proven; not sure \\
 & Epistemic non-verbs (Adjectives, Adverbs, Nouns, Personal Attributions) & Unlikely; likely/morelikely; probably; perhaps; maybe; possible; possibility; seemingly; likelihood; not likely; plausibility; possibly; potentially; potential; to our knowledge; unclear; according to my view; in my opinion; perhaps; doubt; impression; probably; unclear; apparently; uncertainty; uncertain; apparent; assumption; confident; hypothesis; plausibly \\
 & Modal verbs in the simple present & Can; may; may not; must \\
 & Modal verbs in the conditional mood & Could; would; might; should \\
 \bottomrule
\end{tabular}
\end{table}

\FloatBarrier 

\section*{Appendix D}
\stepcounter{section} 
\appendixnumbering
\begin{figure}[H]
  \centering\includegraphics[width=1.1\columnwidth]{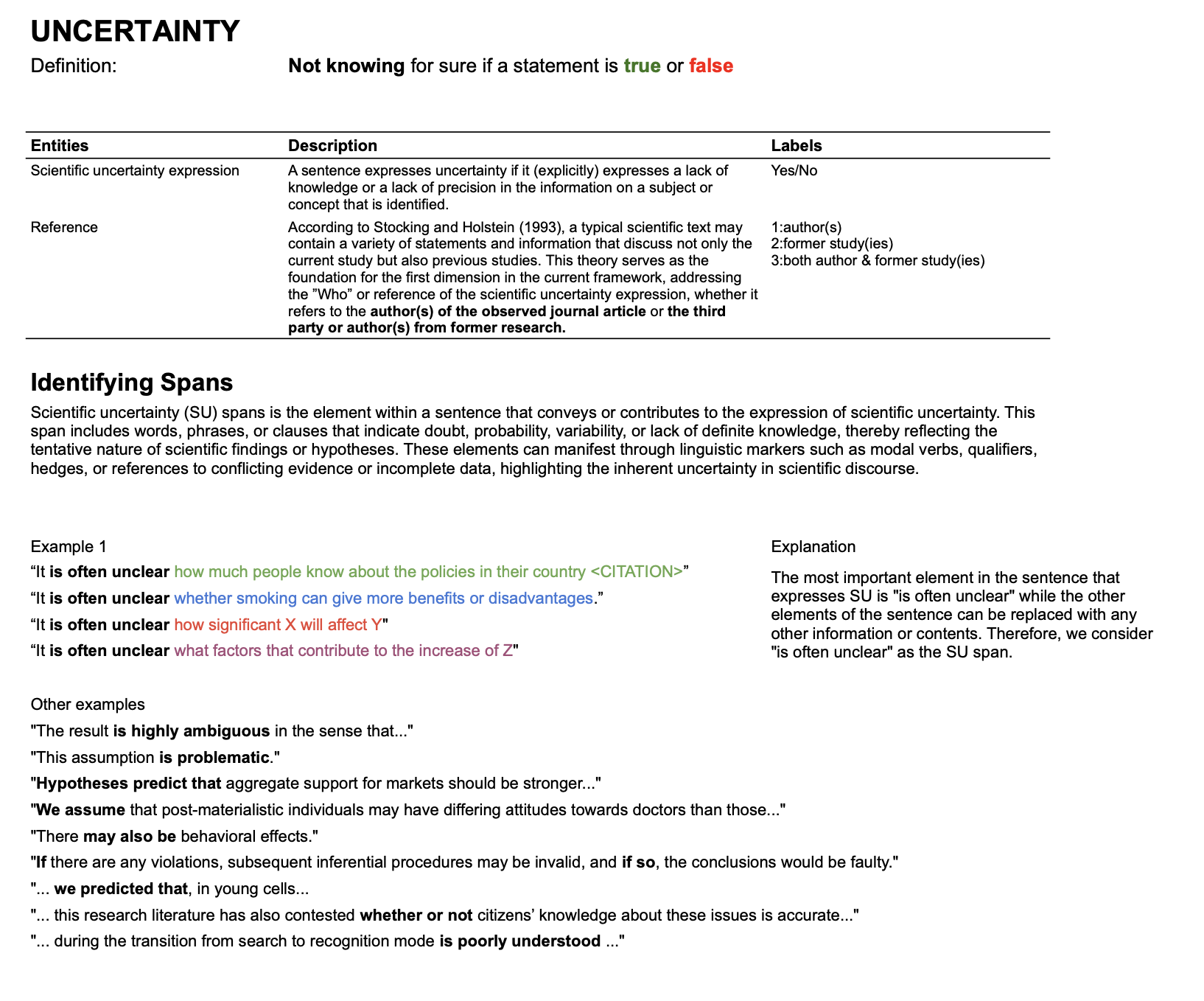}
  \caption{Annotation codebook}
  \label{fig:codebook}
\end{figure}

\FloatBarrier 

\section*{Appendix E}
\stepcounter{section} 
\appendixnumbering
\begin{figure}[H]
  \centering\includegraphics[width=1.1\columnwidth]{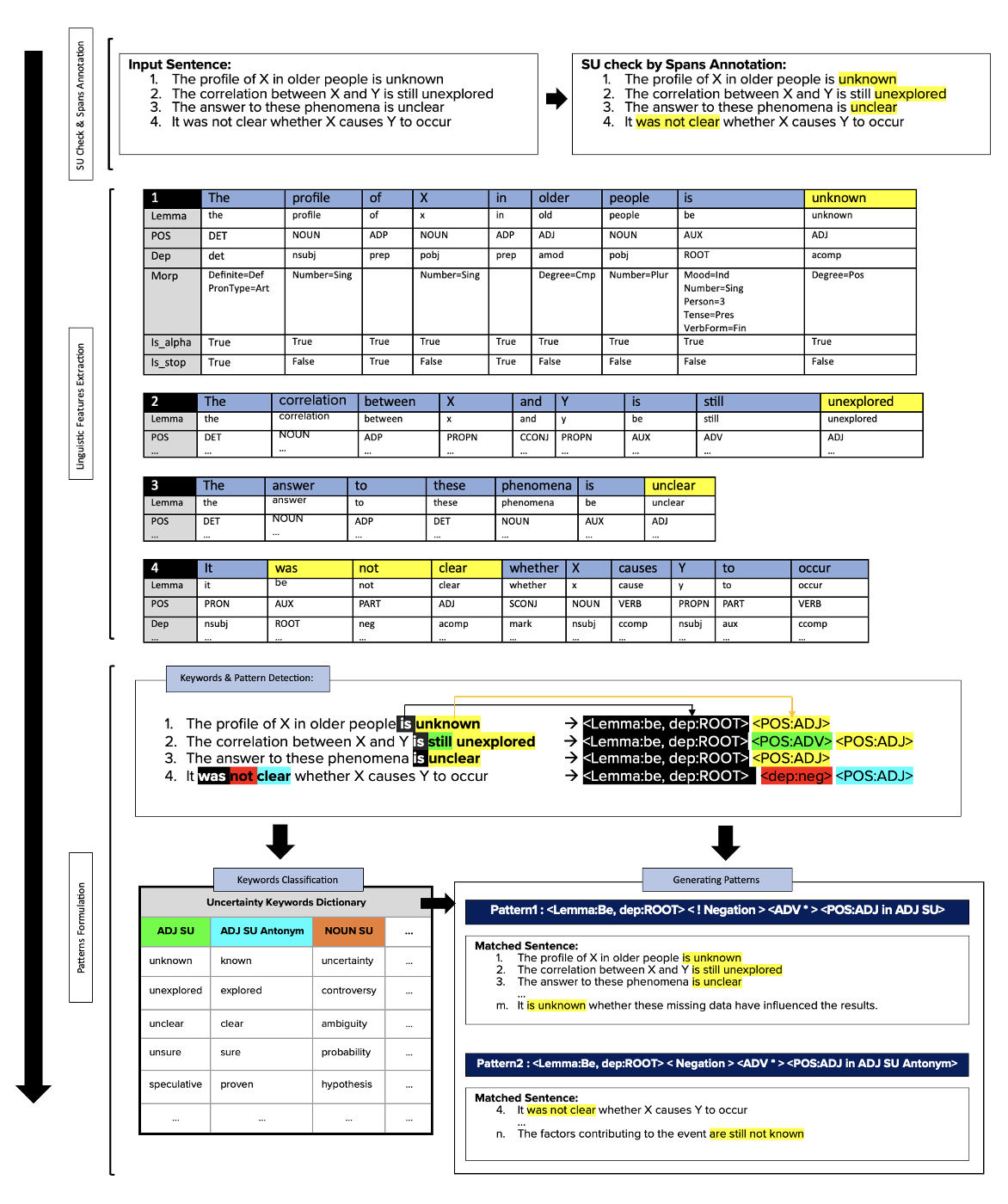}
  \caption{SU patterns formulation}
  \label{fig:patterns_formulation}
\end{figure}


\FloatBarrier 

\section*{Appendix F}
\stepcounter{section} 
\appendixnumbering
\begin{table}[H]
    \centering
   \caption{spaCy model evaluation for extracting lingusitic features from text}
   \label{table:spaCy_models}
    \begin{tabularx}{\textwidth}{lXXXX}
        \toprule
        Model & Accuracy & Precision & Recall & F1 Score \\
        \midrule
        en\_core\_web\_sm & 0.806 & 0.830 & 0.821 & 0.806 \\
        en\_core\_web\_md & 0.807 & 0.831 & 0.822 & 0.807 \\
        en\_core\_web\_lg & 0.805 & 0.829 & 0.820 & 0.805 \\
        en\_core\_web\_trf & 0.808 & 0.832 & 0.823 & 0.808 \\
        \bottomrule
    \end{tabularx}
\end{table}

\end{document}